\documentclass{article} 
\usepackage{arxiv,times}

\usepackage{amsmath,amssymb}
\usepackage{booktabs}
\usepackage{hyperref}
\usepackage{graphicx}
\usepackage{url}
\usepackage{multirow}

\title{Towards Practical Compression of \\ 3D Gaussian Splatting}

\author{%
  Pengpeng~Yu$^{1,2}$, Yueru~Chen$^{2}$, Fei~Song$^{2}$, Tai~Qin$^{3}$, Qi Zhang$^{2,4}$,\\ \textbf{Jing~Wang}$^{2\dagger}$, \textbf{Yulan~Guo}$^{1,2,\dagger}$  \\
  $^{1}$Sun Yat-sen University, China\\
  $^{2}$Pengcheng Laboratory, China\\
  $^{3}${Academy of Broadcasting Science, National Radio and Television Administration, China}\\
  $^{4}$Peking University Shenzhen Graduate School, China\\
  \texttt{yupp5@mail2.sysu.edu.cn}\\
}

\newcommand{\method}{COSA-GS}

\iclrfinalcopy 
\begin{document}

\maketitle

\renewcommand{\thefootnote}{\fnsymbol{footnote}}
\footnotetext[2]{Corresponding author.}
\renewcommand{\thefootnote}{\arabic{footnote}}

\begin{abstract}
3D Gaussian Splatting (3DGS) enables high-quality novel-view synthesis but requires substantial storage.
Existing compression methods often rely on spatial context modeling over irregular 3D representations, increasing the complexity of training and coding. 
Meanwhile, floating-point context inference can introduce numerical inconsistencies across platforms, causing entropy-decoding failures.
To address these practical challenges, we propose \method{}, which constructs \textbf{c}ontext with\textbf{o}ut \textbf{s}patial \textbf{a}ggregation through anchor-wise causal factorization.
Specifically, we use geometry context derived from each anchor's coordinates to model a compact learnable anchor latent.
The anchor latent is then fused with the geometry context to form an anchor context for attribute coding.
The resulting context model features a simple architecture composed solely of linear transformations and activations. 
We train COSA-GS using rate--distortion optimization with adaptive Gaussian pruning. Further, we develop quantization-aware training and integer inference for the context model to achieve bit-exact consistency of entropy-decoded symbols across platforms.
Experiments demonstrate that \method{} achieves state-of-the-art compression performance while retaining fast and consistent cross-platform decoding, providing a simple yet effective framework for practical 3DGS compression. Code is available at \url{https://github.com/pengpeng-yu/COSA-GS}.
\end{abstract}

\section{Introduction}
\label{sec:intro}
3D Gaussian Splatting (3DGS) enables high-quality novel-view synthesis~\citep{kerbl2023_3dgs}.
However, its explicit scene representation often contains numerous Gaussian primitives with associated appearance and geometry parameters, resulting in substantial storage and transmission costs.
To reduce this overhead, existing methods explore pruning, vector quantization, compact parameterizations, and entropy coding~\citep{
bagdasarian2025_3dgszip}.

An important direction is to exploit spatial correlations in the scene representation.
Scaffold-GS~\citep{lu2024_scaffoldgs} parameterizes local Gaussians using anchors and learnable offsets, and predicts their view-dependent attributes from anchor features and viewing conditions, providing a compact representation for subsequent compression methods.
Building on this representation, HAC and HAC++ query a learned hash grid at anchor locations to construct spatial context for attribute distribution modeling~\citep{chen2024_hac,chen2025_hacpp}.
ContextGS instead organizes anchors into hierarchical levels and uses already decoded coarser anchors, together with a per-anchor hyperprior, to predict finer-level attribute distributions~\citep{wang2024_contextgs}.

Although these methods improve compression performance, their reliance on spatial context extraction over irregular 3D structures complicates the coding pipeline and limits coding speed.
Meanwhile, floating-point context inference can introduce numerical inconsistencies across platforms, causing entropy-decoding failures~\citep{balle2019_integer,koyuncu2022_interoperability}.
This raises a practical question: \emph{can we achieve strong rate--distortion performance and fast, consistent cross-platform decoding with a simple architecture?}
Our intuition is that 3DGS optimization itself can reduce inter-anchor dependence when guided by an appropriate context model.
Concretely, we can encourage each anchor to learn distinctive spatial information while concentrating the remaining dependencies within individual anchors, allowing complex spatial context aggregation to be replaced by simpler anchor-wise modeling.
Note that \emph{anchor-wise} does not mean discarding spatial information: the context for each anchor can still be derived from its own coordinate.

Based on this principle, we propose \method{}, short for \textbf{C}ontext with\textbf{o}ut \textbf{S}patial \textbf{A}ggregation for \textbf{G}aussian \textbf{S}platting.
The method organizes context construction and attribute coding into a connected causal structure within each anchor.
Coordinate-derived geometry context is used to model a compact learnable anchor latent, which is then fused with the geometry context to form an anchor context.
The anchor context supports modeling and coding across all attribute groups.
All context modeling uses only anchor-wise linear transformations and activations, yielding a simple compression architecture without spatial context aggregation.
We further develop an integer inference pipeline for the context model, with quantization-aware training to adapt the networks to integer arithmetic.
The pipeline maintains consistency throughout context construction and probability prediction, enabling consistent entropy decoding across platforms.

Our contributions are summarized as follows:
\begin{itemize}
    \item We propose \method{}, a simple 3DGS compression framework that integrates context construction and attribute coding through anchor-wise causal factorization without spatial context aggregation.
    \item We develop an integer inference pipeline with quantization-aware training for the context model, enabling consistent entropy decoding across heterogeneous platforms.
    \item We combine the proposed model with rate--distortion optimization and adaptive masking, achieving state-of-the-art compression performance and fast decoding on Mip-NeRF360, DeepBlending, Tanks\&Temples, and BungeeNeRF.
\end{itemize}

\section{Related Work}
\label{sec:related}

\subsection{3D Gaussian Splatting Compression}

The explicit scene representation of 3DGS enables efficient rendering but incurs substantial storage cost~\citep{kerbl2023_3dgs}.
Recent work has explored compression and compaction through pruning, quantization, compact parameterization, and entropy coding~\citep{
lee2024_compact3dgs,
fan2024_lightgaussian,
niedermayr2024_compressed3dgs,
girish2024_eagles,
papantonakis2024_reducing}.
Post-training and training-free methods reduce storage through pruning and compact attribute coding~\citep{xie2024_mesongs,tian2025_flexgaussian}.
Predictive coding and structured feature-plane representations further reduce redundancy in Gaussian attributes~\citep{liu2024_compgs,lee2025_featureplanes}.
Scene-adaptive lattice vector quantization improves the rate--distortion efficiency of anchor-based compression~\citep{xu2026_salvq}.
Gaussian pruning and sparsification reduce model size by removing redundant primitives~\citep{hanson2025_pup3dgs,zhang2025_gaussianspa}.
Differentiable rate--distortion optimization directly balances storage cost and rendering quality during scene optimization~\citep{wang2024_rdogaussian}.
Anchor-based parameterization represents local Gaussians with anchors and learnable offsets while predicting view-dependent attributes from anchor features and viewing conditions~\citep{lu2024_scaffoldgs}, providing a compact representation for learned 3DGS compression.

\subsection{Context Modeling for 3DGS Compression}
Context modeling improves entropy coding by conditioning symbol probabilities on information available at the decoder, a principle widely used in learned image compression~\citep{balle2018_hyperprior,minnen2018_autoregressive}.
Applying this idea to irregular 3D representations is challenging because neighboring elements do not lie on dense regular grids.
HAC and HAC++ construct spatial context by querying a learned hash grid at anchor locations for anchor-attribute distribution modeling~\citep{chen2024_hac,chen2025_hacpp}.
ContextGS instead builds a coarse-to-fine anchor hierarchy and conditions finer anchors on already decoded coarser anchors, together with a low-dimensional per-anchor hyperprior~\citep{wang2024_contextgs}.
PCGS builds on the HAC++ framework with progressive masking and progressive quantization for incremental bitstreams~\citep{chen2025_pcgs}.
These methods construct context through spatial feature fields or inter-anchor dependencies.
\method{} instead constructs context through causal dependencies among coordinates, latents, and attribute groups within each anchor, retaining conditional context modeling without spatial aggregation.

\subsection{Cross-Platform Decoding in Neural Compression}

Neural compression requires numerical consistency between the encoder and decoder, as even small differences in predicted probabilities can cause entropy-decoding failures.
\citet{balle2019_integer} address platform-dependent floating-point behavior using integer networks for learned image compression.
\citet{koyuncu2022_interoperability} further study cross-device consistency through weight and activation quantization of entropy networks.
For video compression, DCVC-RT incorporates model integerization to support consistent cross-device coding~\citep{jia2025_dcvc_rt}.
Integer inference has also been developed for cross-platform lossless LiDAR point cloud compression~\citep{yu2026_practicalpcc}.
An alternative approach transmits calibration information to correct numerical mismatches at the cost of additional side information~\citep{tian2023_calibration}.
\method{} applies integer inference to the context model of anchor-based 3DGS, maintaining consistency of entropy-decoded symbols across platforms.

\section{Method}
\label{sec:method}

\subsection{Preliminaries}
\label{sec:preliminaries}

Scaffold-GS~\citep{lu2024_scaffoldgs} represents a scene with anchors, each parameterizing $K$ Gaussian primitives.
For anchor $i$, we denote its attributes by
$\mathcal{V}_i=(\mathbf{f}_i,r_i,\mathbf{o}_i,\mathbf{s}_i)$,
where $\mathbf{f}_i\in\mathbb{R}^{C_f}$ is the anchor feature,
$r_i$ is the log-domain position scaling,
$\mathbf{o}_i\in\mathbb{R}^{K\times3}$ contains the Gaussian offsets,
and $\mathbf{s}_i\in\mathbb{R}^{3}$ is the log-domain Gaussian scaling.
Each anchor has a coordinate $\mathbf{x}_i\in\mathbb{R}^{3}$, with the $k$-th Gaussian centered at
$\mathbf{p}_{i,k}=\mathbf{x}_i+\exp(r_i)\mathbf{o}_{i,k}$.
Given a target view, shared MLPs predict view-dependent Gaussian attributes from the anchor feature and viewing conditions:
\begin{equation}
    \left\{
        \left(
        \mathbf{c}_{i,k},
        \alpha_{i,k},
        \mathbf{q}_{i,k},
        \mathbf{u}_{i,k}
        \right)
    \right\}_{k=1}^{K}
    =
    \mathcal{D}_{\theta}(\mathbf{f}_i,\mathbf{v}_i),
    \;\;\;\;
    \mathbf{s}^{\mathrm{G}}_{i,k}
    =
    \exp(\mathbf{s}_i)
    \odot
    \mathrm{sigmoid}(\mathbf{u}_{i,k}),
    \label{eq:scaffold_attributes}
\end{equation}
where $\mathbf{v}_i$ contains the viewing direction and distance, and
$\mathcal{D}_{\theta}$ denotes the shared prediction MLPs and their output transformations.
Here, $\mathbf{c}_{i,k}$, $\alpha_{i,k}$, and $\mathbf{q}_{i,k}$ denote color, opacity, and rotation quaternion, respectively, while $\mathbf{u}_{i,k}$ modulates the base scaling $\exp(\mathbf{s}_i)$.
These attributes define the Gaussian primitives used for rendering.
Scaffold-GS therefore stores anchor attributes and shared MLP weights rather than explicit attributes for every Gaussian.

\begin{figure}[t]
 \centering
 \includegraphics[width=\linewidth,trim=15 129 65 20,clip]{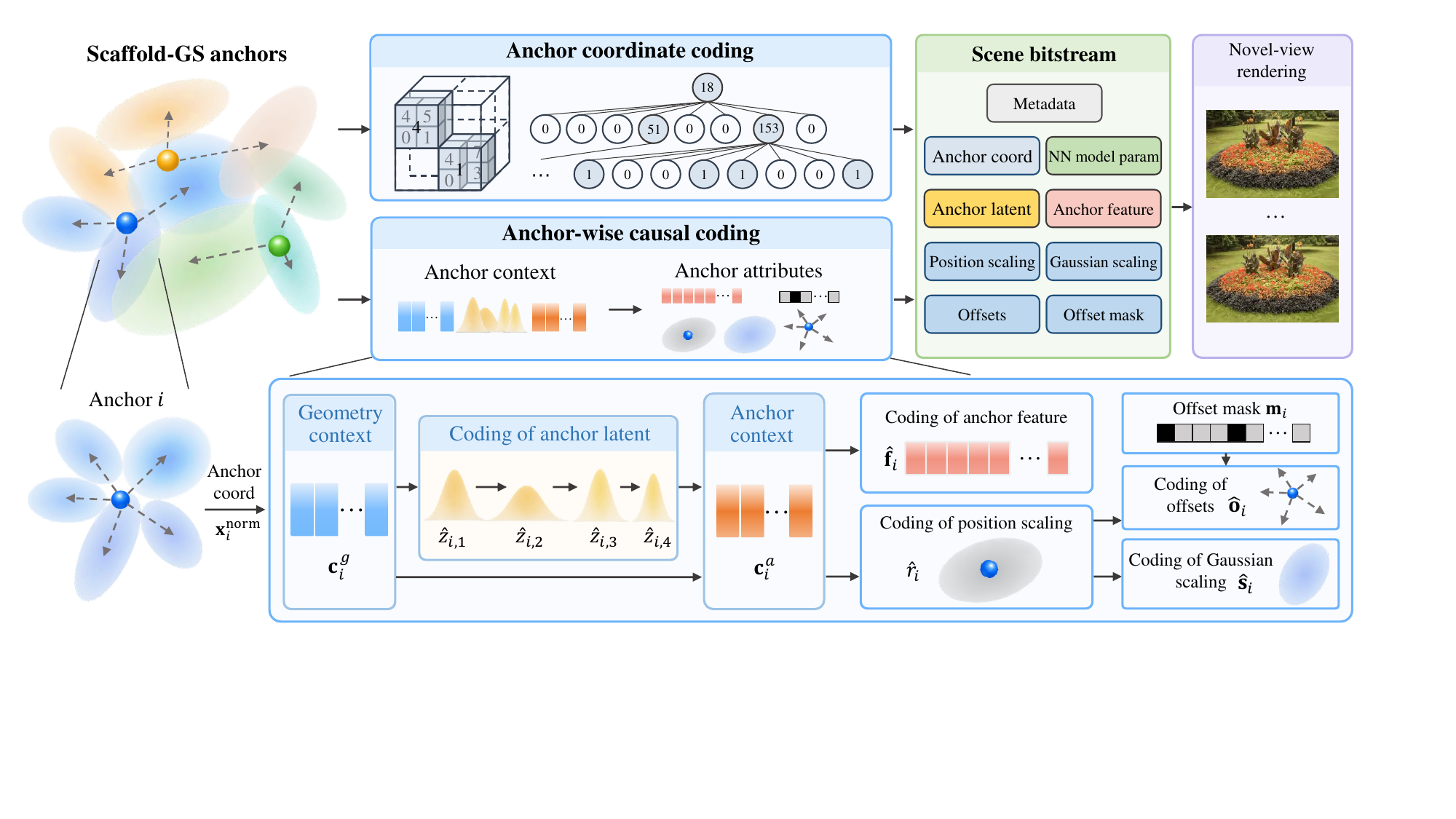}
 \vspace{-1em}
 \caption{Main architecture of \method{}.
The top shows the overall coding and rendering pipeline, while the bottom details the anchor-wise causal structure.
Geometry context supports anchor-latent modeling and is fused with the anchor latent to form the anchor context for attribute coding.}
 \label{fig:overview}
\end{figure}

\subsection{Overview}
\label{sec:method_overview}
As illustrated in Figure~\ref{fig:overview}, \method{} augments the Scaffold-GS representation with learnable anchor latents and compresses it through anchor-wise causal coding.
For a scene with $N$ anchors, we define
$
    \mathcal{A}
    =
    \left\{
        (\mathbf{x}_i,\mathbf{z}_i,\mathcal{V}_i,\mathbf{m}_i)
    \right\}_{i=1}^{N},
$
where $\mathbf{x}_i$ is the anchor coordinate,
$\mathbf{z}_i\in\mathbb{R}^{C_z}$ is the anchor latent,
$\mathcal{V}_i$ denotes the anchor attributes,
and $\mathbf{m}_i\in\{0,1\}^{K}$ is a mask indicating the active Gaussian offsets.

We encode this representation using octree occupancy coding for anchor coordinates and anchor-wise causal coding for anchor latents and attributes.
The anchor latent is modeled using coordinate-derived geometry context and then fused with this context to form the anchor context.
The anchor context supports modeling and coding across all attribute groups.
The resulting bitstream is decoded to recover the Scaffold-GS representation for novel-view rendering. 
Further, quantization-aware training and integer inference for the context model enable consistent cross-platform decoding of entropy-coded symbols.

\subsection{Anchor Coordinate Coding}
We encode anchor coordinates with octree occupancy coding based on empirical distributions.
The coordinates are first quantized as $\mathbf{q}_i=\lfloor \mathbf{x}_i/\Delta_x \rceil$, where $\Delta_x$ is the coordinate quantization step and $\lfloor\cdot\rceil$ denotes rounding.
We construct the octree by recursively grouping occupied cells into their parent cells.
For parent $j$ at level $\ell$, $\omega_{\ell,j}\in\{1,\ldots,255\}$ denotes the occupancy pattern of its eight children.
Occupancy symbols at the same level share an empirical distribution:
\begin{equation}
    p_\ell(v)
    =
    \frac{1}{N_\ell}
    \sum_{j=1}^{N_\ell}
    \mathbf{1}[\omega_{\ell,j}=v],
    \label{eq:occupancy_distribution}
\end{equation}
where $N_\ell$ denotes the number of occupancy symbols at level $\ell$.

During decoding, the octree is expanded level by level to recover the finest-level integer coordinates exactly.
The reconstructed coordinates
$\hat{\mathbf{x}}_i=\Delta_x\mathbf{q}_i$
are subsequently used for geometry context construction.

\subsection{Anchor-Latent Coding and Context Construction}
\label{sec:anchor_context}

We construct the anchor context by modeling the anchor latent from geometry context and then fusing the anchor latent with the geometry context.
For anchor $i$, we first construct a \emph{geometry context}
$\mathbf{c}_i^g=E_x(\mathbf{x}_i^{\mathrm{norm}})$,
where $\mathbf{x}_i^{\mathrm{norm}}$ is obtained by normalizing the reconstructed coordinate $\hat{\mathbf{x}}_i$, and $E_x$ is an MLP.
To enhance this coordinate-derived context, we introduce a compact anchor latent $\mathbf{z}_i$ that is optimized for each scene.
Its joint distribution is factorized as:
\begin{equation}
    p(\hat{\mathbf{z}}_i\mid\mathbf{c}_i^g)
    =
    \prod_{c=1}^{C_z}
    p(\hat z_{i,c}\mid\mathbf{c}_i^g,\hat{\mathbf{z}}_{i,<c}),
    \label{eq:latent_factorization}
\end{equation}
where hats denote quantized reconstructions.
Each latent channel uses a separate MLP to predict its Gaussian mean and standard deviation from the geometry context and previously reconstructed latent channels.
Latent channels are therefore decoded sequentially, while anchors can be processed in parallel.

We use a Gaussian distribution model for anchor latents and attributes.
For a scalar variable $v$ in group $g\in\{z,f,r,o,s\}$, the predicted mean $\mu$ and a learned quantization step $\Delta_g>0$ determine its integer residual and reconstruction, while the predicted standard deviation $\sigma$ determines its coding probability:
\begin{equation}
    k_v =
    \left\lfloor {(v-\mu)}/\Delta_g \right\rceil,
    \qquad
    \hat v=\mu+\Delta_g k_v,
    \qquad
    \ell_v=-\log_2 P(k_v;\sigma),
    \label{eq:gaussian_coding}
\end{equation}
where $P(k_v;\sigma)$ denotes the probability assigned to the residual symbol by a discretized zero-mean Gaussian with standard deviation $\sigma$, and $\ell_v$ is the estimated coding cost.
The quantization step $\Delta_g$ is shared across anchors and components within each group.
During training, rounding is replaced by additive uniform noise, yielding relaxed anchor latents $\tilde{\mathbf{z}}_i$ for subsequent channel prediction and anchor-context construction.

After all latent channels are reconstructed, we construct the \emph{anchor context}:
\begin{equation}
    \mathbf{c}_i^a
    =
    F_a \left(
        [\mathbf{c}_i^g;E_z(\hat{\mathbf{z}}_i)]
    \right),
    \label{eq:anchor_context}
\end{equation}
where $[\cdot;\cdot]$ denotes concatenation, and $E_z$ and $F_a$ are linear transformations with GELU activations.
The resulting anchor context combines coordinate-derived and learned scene-specific information for attribute distribution modeling.

\subsection{Anchor-Attribute Coding}
\label{sec:attribute_coding}

Attribute coding uses the anchor context for all attribute groups.
For anchor $i$, we denote the attributes by
$\mathcal{V}_i=(\mathbf{f}_i,r_i,\mathbf{o}_i,\mathbf{s}_i)$,
representing the anchor feature, position scaling, offsets, and Gaussian scaling, respectively.
Their conditional distribution is factorized as
\begin{equation}
    p(\hat{\mathcal{V}}_i\mid\mathbf{c}_i^a,\mathbf{m}_i)
    = 
    p(\hat{\mathbf{f}}_i\mid\mathbf{c}_i^a)\, \;
    p(\hat r_i\mid\mathbf{c}_i^a)\, \;
    p(\hat{\mathbf{o}}_i\mid\mathbf{c}_i^a,\hat r_i,\mathbf{m}_i)\, \;
    p(\hat{\mathbf{s}}_i\mid\mathbf{c}_i^a,\hat r_i),
    \label{eq:attribute_factorization}
\end{equation}
where $\mathbf{m}_i\in\{0,1\}^{K}$ is a binary offset mask obtained from learnable logits $\boldsymbol{\eta}_i$ as
$\mathbf{m}_i=\mathbf{1}[\mathrm{sigmoid}(\boldsymbol{\eta}_i)>\epsilon_m]$.
Here, $\epsilon_m$ is the masking threshold, and straight-through gradients are used for binarization during training. 
The feature distribution $p(\hat{\mathbf{f}}_i\mid\mathbf{c}_i^a)$ and position-scaling distribution $p(\hat r_i\mid\mathbf{c}_i^a)$ are modeled directly from the anchor context $\mathbf{c}_i^a$ using separate MLPs, with all feature channels predicted in parallel.
After reconstructing $\hat r_i$, we embed it and process the embedding with two branch-specific MLPs, and the resulting features are added to $\mathbf{c}_i^a$ to model the offset distribution
$p(\hat{\mathbf{o}}_i\mid\mathbf{c}_i^a,\hat r_i,\mathbf{m}_i)$
and Gaussian-scaling distribution
$p(\hat{\mathbf{s}}_i\mid\mathbf{c}_i^a,\hat r_i)$. 
The mask $\mathbf{m}_i$ specifies the active offsets, so that only active offsets are modeled and included in the bitstream.

All attribute groups follow the Gaussian distribution modeling in Eq.~\eqref{eq:gaussian_coding}, with a separate quantization step for each group.
In short, context construction and attribute coding form a connected causal structure within each anchor.

\subsection{Rate--Distortion Optimization}
\label{sec:rd_bitstream}

We optimize COSA-GS under a rate--distortion objective that balances rendering quality and the estimated coding rates of anchor latents and attributes. 
For anchor $i$, let $\ell^g_{i,j}$ denote the coding cost of the $j$-th scalar element of group
$g\in\{z,f,r,s\}$, and let $\ell^o_{i,k}$ denote the coding cost of the $k$-th offset.
Here, $m_{i,k}$ is the $k$-th element of the offset mask $\mathbf{m}_i$, and
$a_i=\mathbf{1}[\sum_k m_{i,k}>0]$ indicates whether anchor $i$ is active.
The corresponding rate terms are
\begin{equation}
    R_g = \sum_i a_i\sum_j\ell^g_{i,j},
    \;\;\; g\in\{z,f,r,s\},
    \qquad
    R_o = \sum_i\sum_k m_{i,k}\ell^o_{i,k},
    \label{eq:component_rates}
\end{equation}
where the anchor-level rate terms $R_g$ are applied only to active anchors, and the offset rate $R_o$ is applied only to active offsets. 
The mask $m_{i}$ suppresses inactive Gaussians during rendering, and anchors with no active offsets are pruned.
The total training rate is
$\mathcal{L}_{\mathrm{rate}}=R_z+R_f+R_r+R_o+R_s$.

For a training view with reference image $I$ and rendered image $\tilde I$, the training objective is
\begin{equation}
    \mathcal{L}
    =
    (1-\alpha)\mathcal{L}_1(\tilde I,I)
    +\alpha\bigl(1-\mathrm{SSIM}(\tilde I,I)\bigr)
    + \lambda  \mathcal{L}_{\mathrm{rate}} / (8 \cdot 10^6) ,
    \label{eq:rd_loss}
\end{equation}
where $\mathcal{L}_1$ denotes the mean absolute image error, and $\alpha=0.2$ weights the SSIM term.
The parameter $\lambda$ controls the rate--distortion trade-off.

\subsection{Integer Inference for Cross-Platform Decoding}
\label{sec:integer_entropy}

Cross-platform entropy decoding requires consistent context modeling at the encoder and decoder. 
However, floating-point computations can produce platform-dependent numerical differences and lead to decoding failures.
We therefore implement the context model entirely in integer arithmetic to ensure bit-exact consistency of entropy-decoded symbols across platforms.

\paragraph{Quantization.}
We quantize activations using per-tensor affine quantization and weights using per-channel symmetric quantization, both with 8-bit representations.
For an activation tensor $\mathbf{x}$, the quantized representation is
\begin{equation}
    \mathbf{q}_x =
    \mathrm{clip}\left(
        \left\lfloor \mathbf{x}/s_x \right\rceil + z_x,\,
        q_{\min},\,q_{\max}
    \right),
    \label{eq:network_quantization}
\end{equation}
where $s_x$ and $z_x$ are the quantization scale and zero point, and
$q_{\min}$ and $q_{\max}$ denote the quantization bounds.
During training, we apply quantization-aware training to adapt the context model to these integer representations.
The quantized activations are dequantized as
$s_x(\mathbf{q}_x-z_x)$ for forward computation, with straight-through gradients through rounding.
At inference, the corresponding operations are performed directly in integer arithmetic.

\paragraph{Integer linear.}
Linear transformations use 8-bit integer activations and weights with 32-bit integer accumulation:
\begin{equation}
    \mathbf{a}
    =
    \mathbf{Q}_w(\mathbf{q}_x-z_x)
    +\mathbf{q}_b,
    \label{eq:integer_linear}
\end{equation}
where $\mathbf{q}_x$ is the quantized input, $\mathbf{Q}_w$ and $\mathbf{q}_b$ are the quantized weight matrix and bias, $z_x$ is the input zero point, and $\mathbf{a}$ is the accumulated integer output.
The subtraction of $z_x$ is applied element-wise.
Since the weights use symmetric quantization, their zero point is zero.
To convert the accumulated output to the quantization scale of the next layer, we use fixed-point rescaling.
The scale conversion is represented by integer multipliers $\boldsymbol{\rho}$ and a power-of-two divisor $2^\tau$, giving
\begin{equation}
    \mathbf{q}_y
    =
   \mathrm{clip}\left(
    \left\lfloor
    \frac{\mathbf{a} \odot \boldsymbol{\rho}}{2^{\tau}}
    \right\rceil
    + z_y,\,
    q_{\min},\,q_{\max}
\right),
    \label{eq:integer_requantization}
\end{equation}
where $\boldsymbol{\rho}$ contains the precomputed integer multipliers, $\tau$ is the right-shift factor, $z_y$ is the output zero point, and $\odot$ denotes element-wise multiplication.

\paragraph{Integer GELU.}
The GELU activation is
$\mathrm{GELU}(x)=x\Phi(x)$,
where $\Phi$ is the standard Gaussian cumulative distribution function.
To avoid floating-point calculation of $\Phi$, we rewrite GELU as
\begin{equation}
    \mathrm{GELU}(x)
    =
    \max(x,0)-h(|x|),
    \;\;\;\;
    h(t)=t\,\Phi(-t).
    \label{eq:gelu_decomposition}
\end{equation}
This decomposition isolates the nonlinear term into $h(|x|)$.
We approximate $h$ using a quantized lookup table $\mathcal{T}$.
Since $h$ approaches zero for large $|x|$, the lookup table only needs to cover a bounded nonnegative range, while $\max(x,0)$ can be evaluated directly in integer arithmetic.
For a fixed-point input $u$, the integer GELU is computed as
$
    \mathcal{G}(u)
    =
    \max(u,0)-\mathcal{I}_{\mathcal{T}}(|u|),
$
where $\mathcal{I}_{\mathcal{T}}$ denotes integer linear interpolation of the lookup-table values at the corresponding fixed-point scale.
The correction term is set to zero outside the tabulated range.

\paragraph{Integer Gaussian modeling.}
We use precomputed integer cumulative distribution function (CDF) tables to ensure consistent Gaussian entropy coding across platforms.
The tables represent the discretized zero-mean Gaussian distributions in Eq.~\eqref{eq:gaussian_coding}.
The standard-deviation index predicted by the context model is rounded and clipped using integer arithmetic to select the corresponding CDF table.
For reconstruction, the residual symbol $k_v$ is rescaled by $\Delta_g$ and added to the predicted mean $\mu$ in fixed-point arithmetic:
\begin{equation}
    q_{\hat v}
    =
    q_\mu
    +
    \left\lfloor
        (k_v \rho_v) / 2^{\tau_v}
    \right\rceil,
    \label{eq:integer_gaussian_reconstruction}
\end{equation}
where $q_\mu$ and $q_{\hat v}$ are the fixed-point representations of the predicted mean $\mu$ and reconstructed value $\hat v$.
The integer multiplier $\rho_v$ and right-shift factor $\tau_v$ are precomputed from the quantization step $\Delta_g$ and the fixed-point scale of $q_\mu$.
Therefore, CDF selection and reconstruction for subsequent context modeling use integer arithmetic, maintaining consistency across platforms.

\section{Experiments}
\label{sec:experiments}

\subsection{Settings}
\label{sec:experimental_settings}

\paragraph{Datasets and compared methods.}
Experiments are conducted on Mip-NeRF360~\citep{barron2022_mipnerf360}, DeepBlending~\citep{hedman2018_deepblending}, Tanks\&Temples~\citep{knapitsch2017_tandt}, and BungeeNeRF~\citep{xiangli2022_bungeenerf}.
We compare against ContextGS~\citep{wang2024_contextgs}, HAC++~\citep{chen2025_hacpp}, CAT-3DGS~\citep{zhan2025_cat3dgs}, PCGS~\citep{chen2025_pcgs}, and SpeedyGS~\citep{zhang2026_speedygs}.
We report dataset-average PSNR, SSIM, and LPIPS for novel-view rendering, together with compressed scene size in MiB.
Higher PSNR and SSIM and lower LPIPS indicate better novel-view quality. The rate--distortion results of the compared methods are based on the per-scene metrics reported in the corresponding papers.

\paragraph{Implementation details.}
We initialize from a Scaffold-GS representation optimized for 15K iterations and use 30K rate--distortion optimization iterations as the default setting.
The notation 15+30K therefore denotes 45K iterations in total.
We also evaluate 15+20K and 15+45K settings and a 15+30K quantization-aware training (QAT) variant.
Unless otherwise specified, experiments use an NVIDIA 5880 Ada GPU and an AMD EPYC 9654 CPU. 
More implementation details are provided in the appendix and the repository.

\begin{figure}[t]
  \centering
  \includegraphics[width=1.0\linewidth,trim=0 645 35 0,clip]{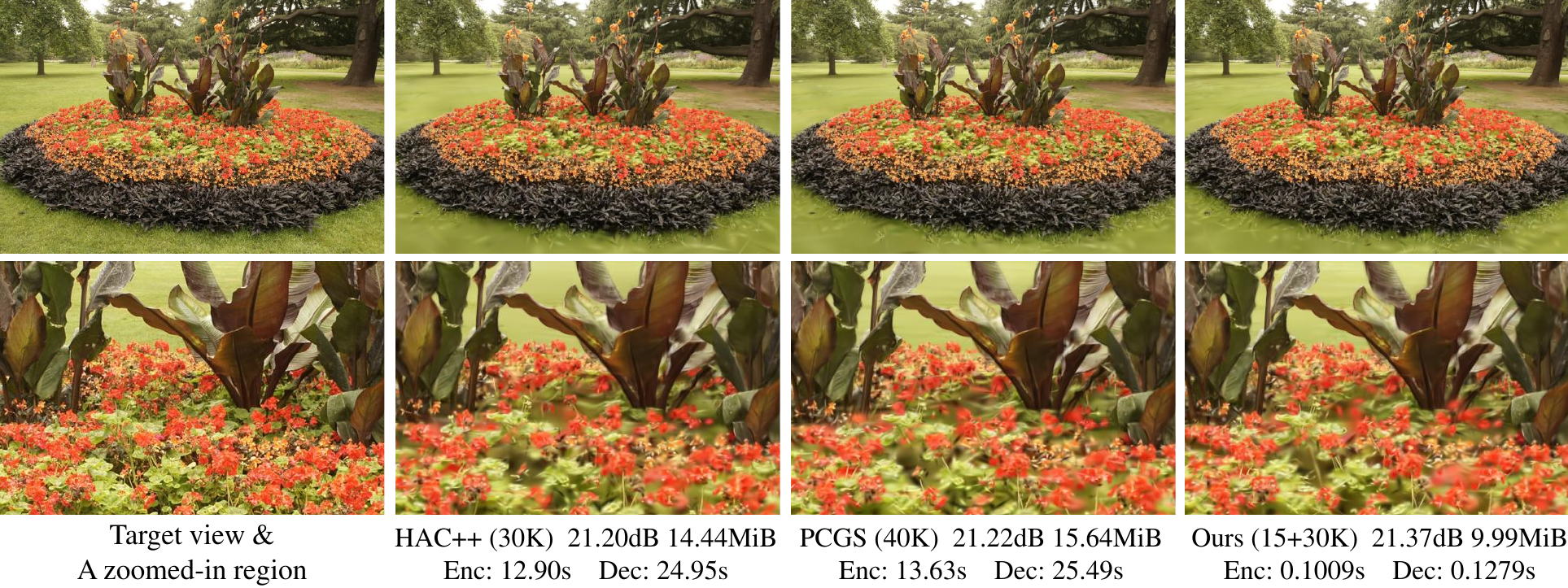}
  \vspace{-1.5em}
  \caption{Qualitative comparison on the \textit{flowers} scene from Mip-NeRF360.
The top row shows the target view \textit{DSC9144} and the corresponding renderings, while the bottom row shows zoomed-in views of a distinctive region. Scene sizes and coding times are reported for the full scene. PSNR is measured on the displayed target view. Coding times are measured on a platform with an NVIDIA 5880 Ada GPU and an AMD EPYC 9654 CPU.}
  \label{fig:vis}
\end{figure}

\begin{figure}[t]
  \centering
  \includegraphics[width=0.328\linewidth,trim=5 6 5 5,clip]{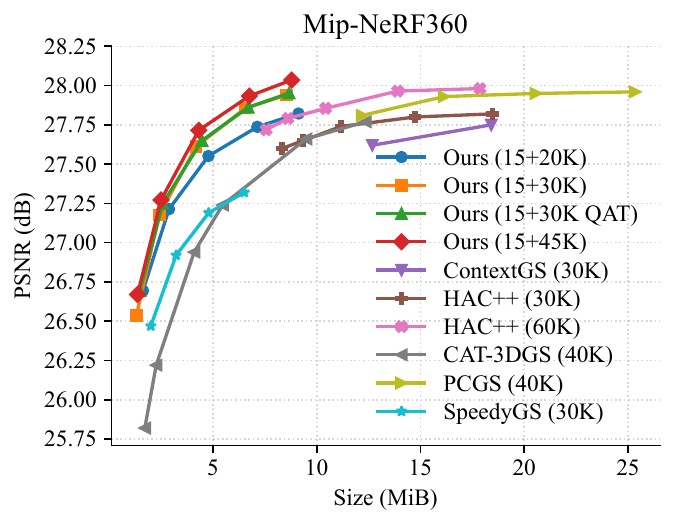}
  \includegraphics[width=0.328\linewidth,trim=5 6 5 5,clip]{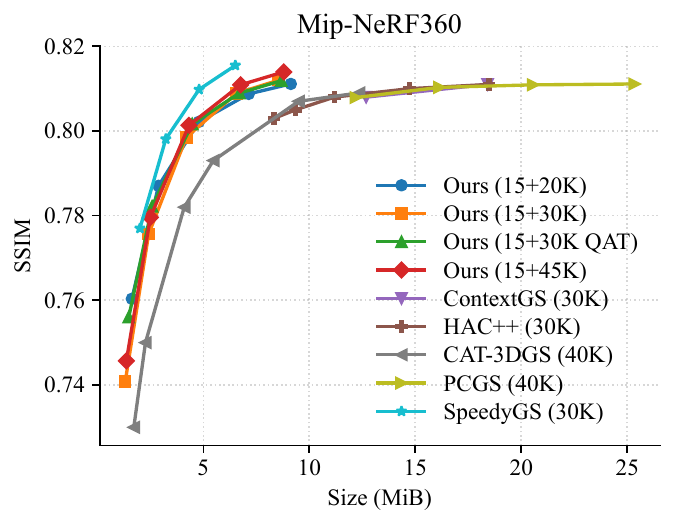}
  \includegraphics[width=0.328\linewidth,trim=5 6 5 5,clip]{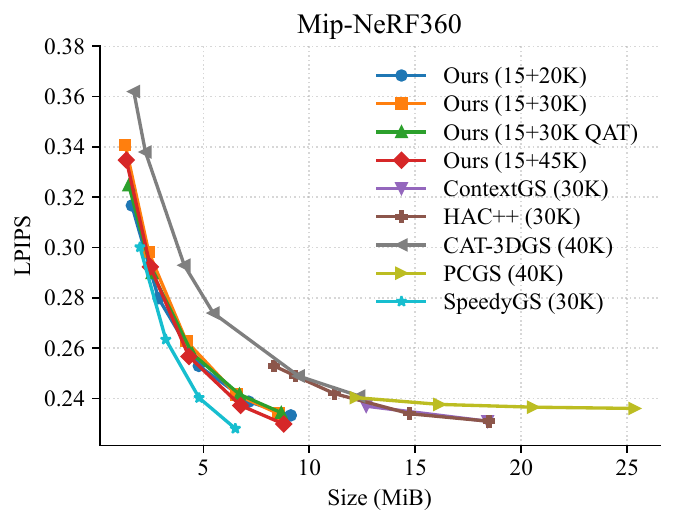}
  \\
  \includegraphics[width=0.328\linewidth,trim=5 6 5 5,clip]{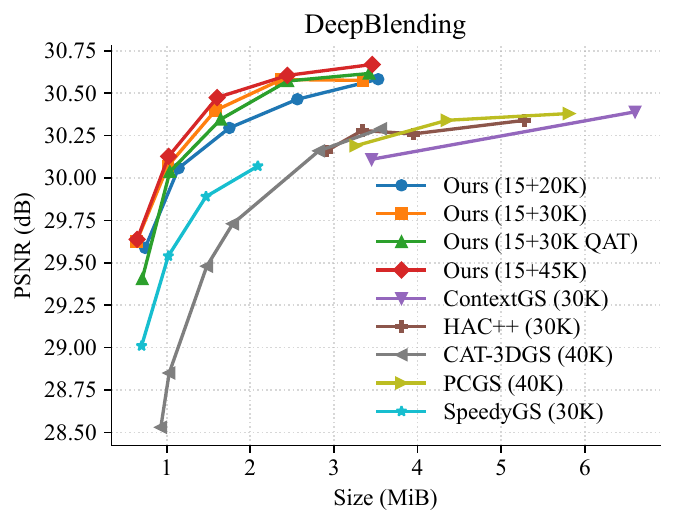}
  \includegraphics[width=0.328\linewidth,trim=5 6 5 5,clip]{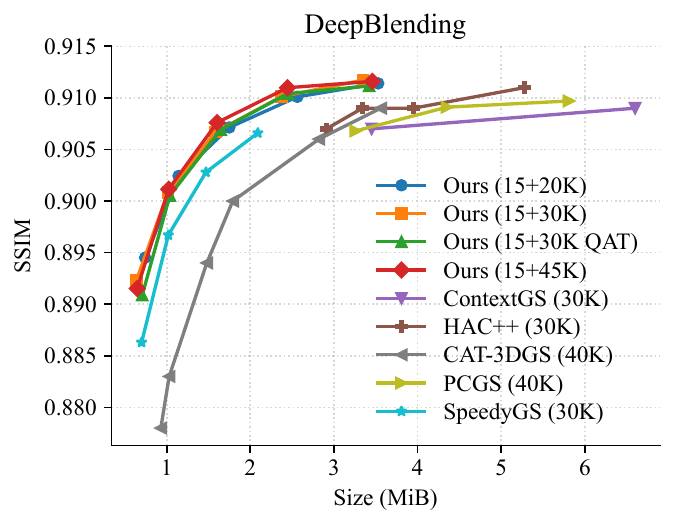}
  \includegraphics[width=0.328\linewidth,trim=5 6 5 5,clip]{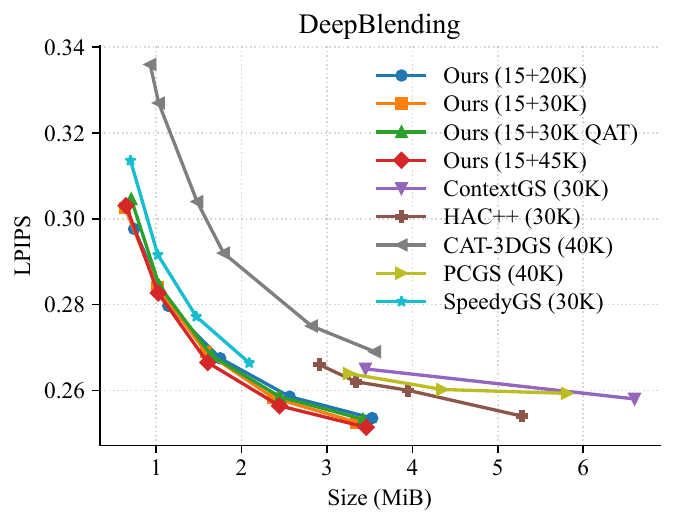}
  \includegraphics[width=0.328\linewidth,trim=5 6 5 5,clip]{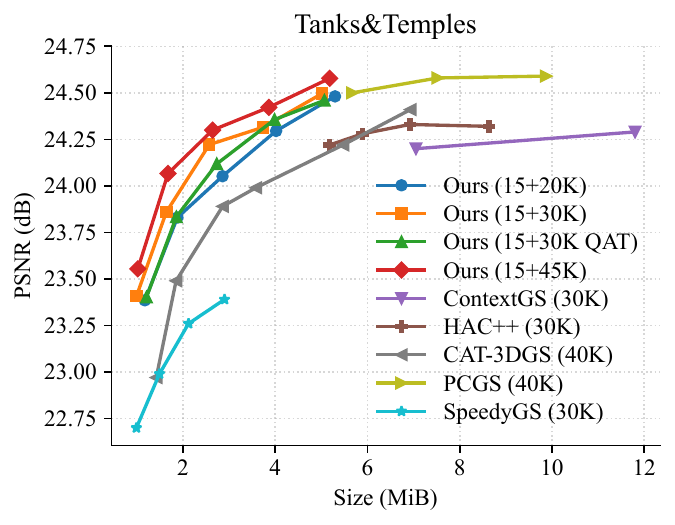}
  \includegraphics[width=0.328\linewidth,trim=5 6 5 5,clip]{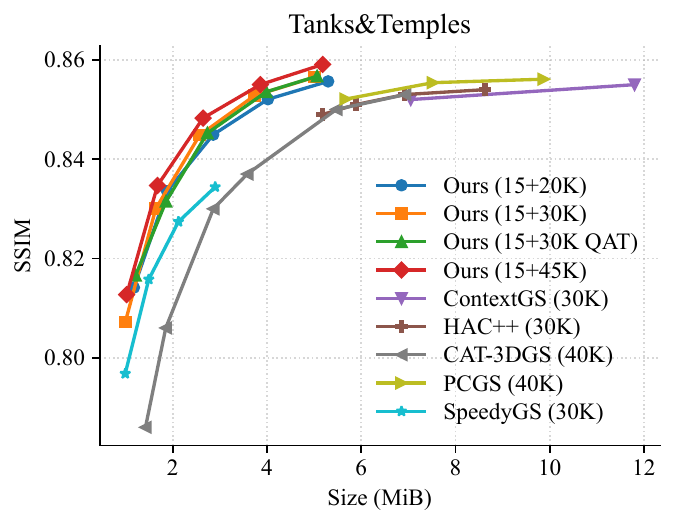}
  \includegraphics[width=0.328\linewidth,trim=5 6 5 5,clip]{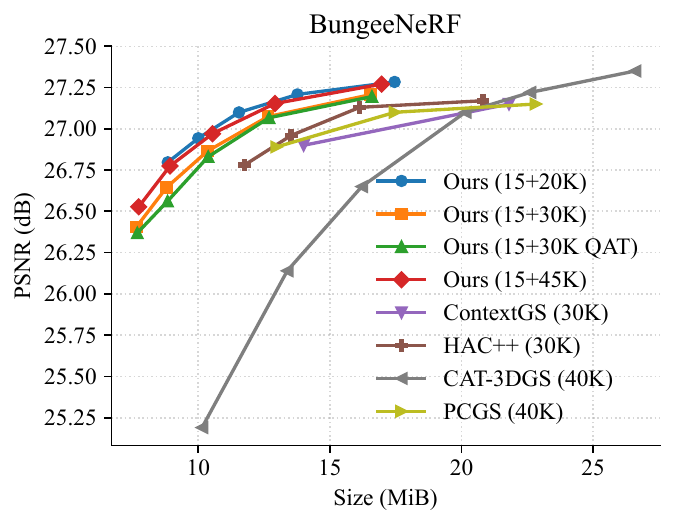}
  \vspace{-1.5em}
  \caption{Rate--distortion comparison with existing methods.
The first two rows show PSNR, SSIM, and LPIPS on Mip-NeRF360 and DeepBlending, respectively.
The last row shows PSNR and SSIM on Tanks\&Temples and PSNR on BungeeNeRF.
Training steps are shown in parentheses, and QAT denotes quantization-aware training.}
  \label{Fig:RD}
\end{figure}

\subsection{Performance Analysis}
\label{sec:performance_analysis}

\paragraph{Rate--distortion performance.}
Figure~\ref{Fig:RD} compares the rate--distortion performance across the four datasets.
We first examine Mip-NeRF360, where the default 15+30K setting of \method{} achieves a favorable PSNR--size trade-off compared with existing methods, including HAC++ with 60K optimization steps.
As reported in Table~\ref{tab:decoding_breakdown}, \method{} covers 1.32--8.55 MiB and 26.54--27.94 dB on Mip-NeRF360, reaching 27.61 dB at 4.20 MiB.
Meanwhile, the SSIM and LPIPS curves also show better overall trade-offs than ContextGS, HAC++, CAT-3DGS, and PCGS at comparable sizes.
SpeedyGS achieves higher SSIM and lower LPIPS at several bitrate points, while \method{} provides a stronger PSNR--size trade-off.
On DeepBlending, Tanks\&Temples, and BungeeNeRF, the default 15+30K setting similarly achieves favorable PSNR with smaller scene sizes than the compared methods.
On the \textit{flowers} scene in Figure~\ref{fig:vis}, \method{} achieves a decoding time of 0.1279 s, compared with 24.95 s for HAC++ and 25.49 s for PCGS, while producing a smaller bitstream and slightly better rendering quality.

Figure~\ref{Fig:RD} also compares different numbers of optimization steps.
Increasing the optimization steps generally improves rate--distortion performance, but the gain gradually diminishes and varies across datasets.
The improvement is more evident on Mip-NeRF360, DeepBlending, and Tanks\&Temples, whereas the curves on BungeeNeRF remain close across the 15+20K, 15+30K, and 15+45K settings.
The QAT variant applies quantization-aware training together with the integer context-model inference described in Sec.~\ref{sec:integer_entropy}.
Its rate--distortion curves remain close to the floating-point 15+30K model over most bitrate points, with only small differences at some low-bitrate points on DeepBlending and BungeeNeRF.
Overall, these results show that the proposed anchor-wise context modeling is effective across the four benchmarks.

\begin{table}[tb]
  \centering
  \vspace{-1em}
  \footnotesize
  \caption{Novel-view quality, size, and runtime of COSA-GS (15+30K) on Mip-NeRF360.}
  \vspace{0.5em}
  \renewcommand{\arraystretch}{0.95}
  \setlength{\tabcolsep}{2.4pt}
  \label{tab:decoding_breakdown}
    \begin{tabular}{cccccccccccc}
    \toprule
    \multirow{2}[2]{*}{$\lambda$} & \multirow{2}[2]{*}{PSNR} & \multirow{2}[2]{*}{SSIM} & \multirow{2}[2]{*}{LPIPS} & Size & Total & Total & Entropy & {\fontsize{8pt}{9.6pt}\selectfont Anchor Coord.} & NN  & \multicolumn{2}{c}{Train (s)} \\
        &     &     &     &  (MiB) & Enc. (s) & Dec. (s) &  Dec. (s) & Dec. (s) & Dec. (s) & {\fontsize{8pt}{9.6pt}\selectfont Scaffold15K} & {\fontsize{8pt}{9.6pt}\selectfont RD 30K} \\
    \midrule
    .0006 & 27.94 & 0.812 & 0.234 & 8.55 & 0.090 & 0.112 & 0.090 & 0.017 & 0.008 & \multirow{5}[2]{*}{479} & 3321 \\
    .0010 & 27.86 & 0.809 & 0.242 & 6.56 & 0.080 & 0.101 & 0.082 & 0.015 & 0.007 &     & 3279 \\
    .0020 & 27.61 & 0.798 & 0.263 & 4.20 & 0.064 & 0.081 & 0.065 & 0.012 & 0.006 &     & 3238 \\
    .0040 & 27.18 & 0.776 & 0.298 & 2.42 & 0.048 & 0.062 & 0.048 & 0.010 & 0.004 &     & 3206 \\
    .0080 & 26.54 & 0.741 & 0.341 & 1.32 & 0.036 & 0.044 & 0.034 & 0.007 & 0.003 &     & 3210 \\
    \midrule
    Avg. & 27.42 & 0.787 & 0.276 & 4.61 & 0.063 & 0.080 & 0.064 & 0.012 & 0.005 & 479 & 3251 \\
    \bottomrule
    \end{tabular}%
\end{table}

\begin{table}[t]
  \centering
  \footnotesize
  \vspace{-1em}
  \renewcommand{\arraystretch}{0.95}
  \setlength{\tabcolsep}{2.8pt}
  \caption{Cross-platform decoding times of COSA-GS (15+30K, QAT) on Mip-NeRF360. All bitstreams are encoded on the NVIDIA 5880 Ada and AMD EPYC 9654 platform and decoded on the listed GPU--CPU configurations. Times are in seconds. The final column reports the number of bitstreams with identical entropy-decoded symbols and anchor coordinates out of 45 tested bitstreams.}
  \vspace{0.5em}
  \label{tab:cross_platform}
    \begin{tabular}{llccccccc}
    \toprule
    \multicolumn{1}{c}{\multirow{2}[4]{*}{GPU}} & \multicolumn{1}{c}{\multirow{2}[4]{*}{CPU}} & \multicolumn{6}{c}{Total decoding time (s)} & Decoding \\
\cmidrule{3-8}        &     & $\lambda=.0006$ & $.001$ & $.002$ & $.004$ & $.008$ & Average & consistency \\
    \midrule
    NVIDIA 5880Ada & AMD EPYC 9654 & 0.121 & 0.110 & 0.096 & 0.076 & 0.061 & 0.093 & 45/45 \\
    NVIDIA 4090 & AMD EPYC 7R32 & 0.145 & 0.134 & 0.114 & 0.090 & 0.071 & 0.111 & 45/45 \\
    NVIDIA V100 & Intel Xeon Gold 6258R & 0.184 & 0.173 & 0.149 & 0.120 & 0.096 & 0.144 & 45/45 \\
    \bottomrule
    \end{tabular}%
\end{table}

\paragraph{Computational efficiency.}
Table~\ref{tab:decoding_breakdown} reports novel-view quality, scene size, runtime, and training time on Mip-NeRF360. 
Total decoding takes 0.044--0.112 s across the five bitrate points, averaging 0.080 s.
The entropy coder is the dominant decoding component, taking 0.064 s on average, whereas anchor-coordinate decoding takes only 0.012 s.
The neural context model requires only 0.003--0.008 s, indicating that the anchor-wise context modeling introduces little decoding overhead. 
Note that the sum of the component decoding times may slightly exceed the total decoding time, as expected with asynchronous GPU execution.
For training, the shared 15K Scaffold-GS initialization takes 479 s, while the subsequent 30K rate--distortion optimization takes 3,206--3,321 s across bitrate points and accounts for most of the training time.

\paragraph{Cross-platform decoding.}
We encode the scene bitstreams on the NVIDIA 5880 Ada and AMD EPYC 9654 platform and decode the same streams on the three GPU--CPU configurations in Table~\ref{tab:cross_platform}.
On the reference 5880 Ada platform, dataset-average decoding takes 0.061--0.121 s across the five bitrate points, averaging 0.093 s.
The corresponding ranges are 0.071--0.145 s on the NVIDIA 4090 configuration and 0.096--0.184 s on the NVIDIA V100 configuration.
Across all three platforms, decoding time decreases toward lower bitrate points, reflecting the reduced amount of coded data.
We verify cross-platform consistency by checking exact equality of the entropy-decoded symbols for anchor latents, anchor attributes, offset masks, and the decoded anchor coordinates.
All bitstreams of the 5 bitrate points on the 9 Mip-NeRF360 scenes are decoded consistently across platforms.

\begin{figure}[t]
  \centering
  \includegraphics[width=0.328\linewidth,trim=5 6 5 5,clip]{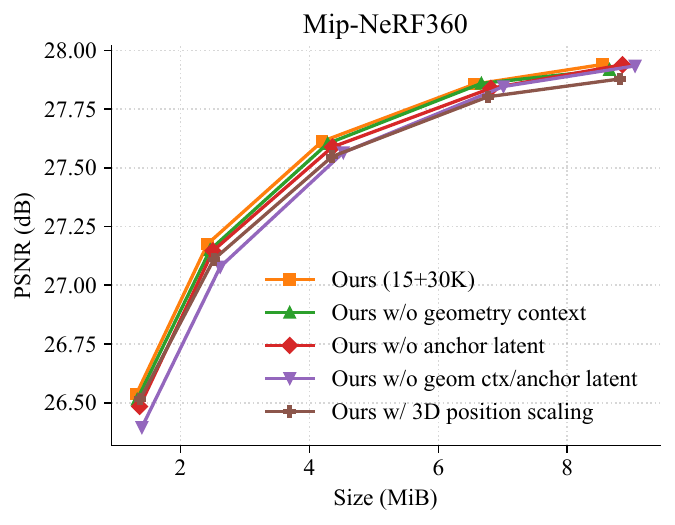}
  \includegraphics[width=0.328\linewidth,trim=5 6 5 5,clip]{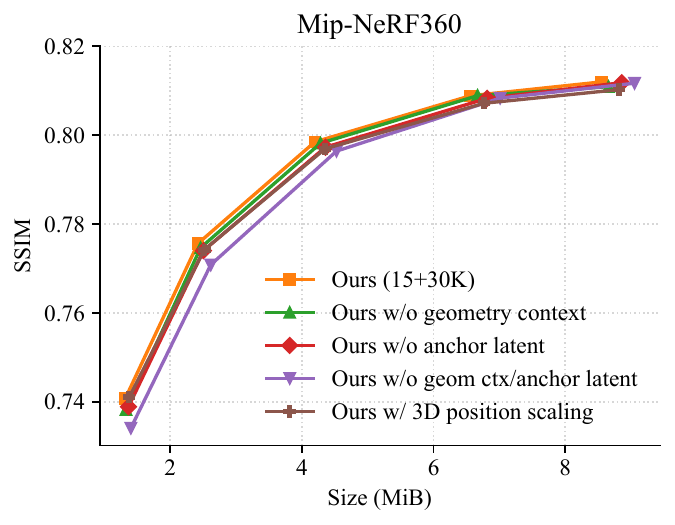}
  \includegraphics[width=0.328\linewidth,trim=5 6 5 5,clip]{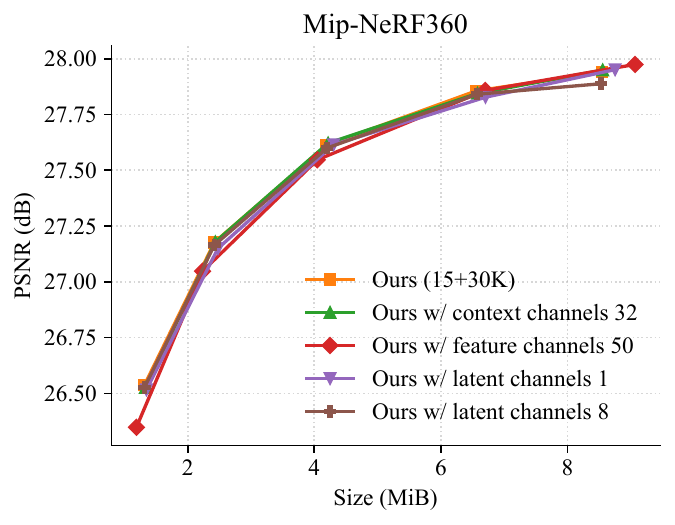}
  \caption{Ablation studies on Mip-NeRF360 under the 15+30K setting. Left and middle: the effects of geometry context, anchor latents, and position-scaling dimensionality. Right: comparisons of different numbers of context, feature, and latent channels.}
  \label{Fig:RD_Ablation}
\end{figure}

\subsection{Ablation Studies}
\label{sec:ablation_studies}

We conduct ablations on Mip-NeRF360 using the 15+30K setting to validate
context construction, position-scaling dimensionality, and model capacity. 

\paragraph{Geometry context and anchor latent.}
Figure~\ref{Fig:RD_Ablation} compares the full model with variants that remove the geometry context, the anchor latent, or both.
Removing either source causes a small degradation, indicating that the two context sources provide complementary information.
When both are removed, performance drops noticeably across all bitrate points, showing that anchor context is essential to the compression performance of \method{}.
The geometry-free variant remains close to the full model, further indicating that the learnable anchor latent can provide informative context even without coordinate conditioning.

\paragraph{Position scaling.}
COSA-GS uses a scalar position scaling for each anchor to scale its offsets uniformly.
To examine whether additional directional flexibility benefits compression, we replace the scalar with a three-dimensional vector.
Figure~\ref{Fig:RD_Ablation} shows that this variant provides no overall rate--distortion improvement and slightly degrades performance at high bitrate points.
The additional scaling degrees of freedom therefore increase the amount of scaling information to be coded without providing sufficient quality gains to offset the added rate.

\paragraph{Model capacity.}
The right panel of Figure~\ref{Fig:RD_Ablation} evaluates the number of channels used for the context, anchor feature, and anchor latent.
Increasing the context channels from 24 to 32 provides little improvement.
Increasing the feature channels from 32 to 50 slightly improves PSNR at the highest bitrate point but degrades performance at lower bitrates.
Changing the number of latent channels from the default four to either one or eight also yields similar rate--distortion curves, with no consistent gain from using more latent channels.
Overall, increasing the number of channels provides limited benefit, supporting the default configuration of COSA-GS.

\section{Conclusion}
We presented \method{}, an anchor-wise causal framework for compact 3DGS compression without spatial context aggregation.
A compact anchor latent is modeled from coordinate-derived geometry context and then fused with this context to provide context for attribute coding, yielding a simple context model composed only of linear transformations and activations.
Combined with rate--distortion optimization, quantization-aware training, and integer inference, \method{} achieves strong compression performance, fast decoding, and cross-platform consistency.
Experiments across four benchmarks demonstrate the effectiveness of this simple anchor-wise architecture for practical 3DGS compression.

\subsection*{AI use statement}

In this work, we used generative AI tools to assist with language editing and the
implementation and debugging of code. We have reviewed the revised
text and inspected and tested the AI-assisted code. AI assistance was
limited to these tasks. We take responsibility for the
manuscript, code, and reported results.

\subsection*{Reproducibility statement}
We provide source code at \url{https://github.com/pengpeng-yu/COSA-GS}, including model implementations, training configurations, and scripts for compression, decompression, and evaluation. 
The accompanying README provides environment setup instructions and commands for running the experiments. 
We encourage readers to inspect the implementation and reproduce the experiments. 


\bibliography{iclr2027_conference}
\bibliographystyle{iclr2027_conference}

\newpage
\appendix
\section{Appendix}
\label{sec:appendix}

This appendix provides implementation and runtime details, bit allocation analysis, extended rate--distortion comparisons, additional ablation studies, and detailed results for individual scenes.

\subsection{Dataset configuration}
We follow the image-resolution settings and training/test view splits used in the publicly released code of Scaffold-GS, HAC++, PCGS, and CAT-3DGS.
We evaluate on nine scenes from Mip-NeRF360, two from Tanks\&Temples, two from DeepBlending, and six from BungeeNeRF.
Images wider than 1,600 pixels are downsampled to a width of 1,600 while preserving their aspect ratio, and smaller images retain their original resolution.
Views are sorted by image filename before splitting.
For Mip-NeRF360, Tanks\&Temples, and DeepBlending, every eighth view, starting from the first, is held out for testing, and the remaining views are used for training.
For BungeeNeRF, the first 31 views are used for testing and the remaining views for training.

\subsection{Bit allocation.}
\begin{figure}[ht]
  \centering
  \includegraphics[scale=0.66,trim=20 6 225 20,clip]{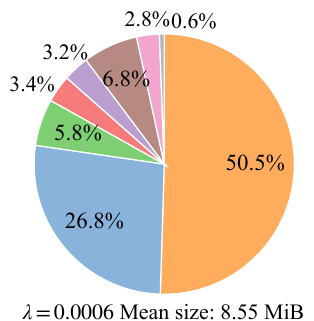}
  \includegraphics[scale=0.66,trim=15 6 227 20,clip]{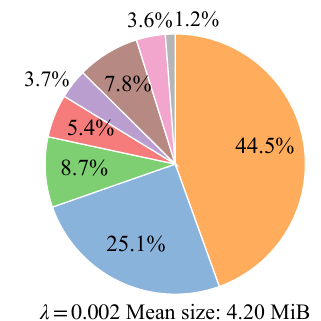}
  \includegraphics[scale=0.66,trim=13 6 225 20,clip]{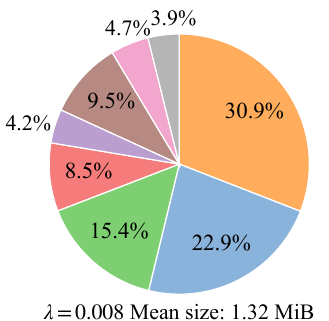}
  \includegraphics[scale=0.66,trim=184 13 90 20,clip]{fig/plot_bitrate_breakdown/mipnerf360_ours_30k_lmb0.008.pdf}
  \vspace{-0.5em}
  \caption{Bitstream composition on Mip-NeRF360 for $\lambda=0.0006$, $0.002$, and $0.008$ under the 15+30K setting. The corresponding mean scene sizes are 8.55, 4.20, and 1.32 MiB, respectively.}
  \label{Fig:BitrateBreakdown}
\end{figure}

Figure~\ref{Fig:BitrateBreakdown} shows the bitstream composition at three rate settings on Mip-NeRF360.
Anchor features and offsets account for most of the bitstream, contributing 77.3\% at $\lambda=0.0006$ and 53.8\% at $\lambda=0.008$.
Anchor latents contribute 3.4\%--8.5\%, providing context for multiple attribute distributions with a comparatively small cost.
As the total rate decreases, the fraction of anchor coordinates increases from 5.8\% to 15.4\%, while the fraction of model parameters and metadata increases from 0.6\% to 3.9\%.
The low-rate bitstreams therefore allocate a larger fraction to anchor coordinates, model parameters, and metadata as the coding costs of features and offsets decrease.

\subsection{Runtime measurement}
\label{app:runtime-measurement}

\paragraph{Measurement protocol.}
We measure encoding and decoding times after one warmup run in the same process.
For encoding--decoding evaluation, we first encode the trained model and decode the resulting bitstream as a warmup, then repeat the complete procedure and record the encoding and decoding times separately.
For decoding-only evaluation, we decode the same bitstream once as a warmup and measure the second decoding run.
Warmup is excluded from the reported times.

\paragraph{Timing boundaries.}
We use a host wall clock and synchronize CUDA immediately before and after each measured encoding or decoding call.
Encoding starts from a trained model in memory and ends with an in-memory scene bitstream.
Decoding starts from that bitstream and an initialized decoder model, and ends with the coordinates, mask, attributes, and network parameters required for rendering.
Both intervals include context prediction, entropy coding, CPU--GPU transfers, and in-memory serialization or parsing.
Checkpoint loading, scene-file input/output, decoder-model construction, rendering, and image-quality evaluation are excluded.
For the QAT variant, encoding time includes exporting the quantized networks. Decoding time excludes constructing empty integer-network modules but includes restoring and transferring their parameters.

\paragraph{Entropy coding and component times.}
We use a CPU rANS backend with 16-bit probability precision.
Each Gaussian-coded stream is split into four independent blocks by default and processed by up to four CPU worker threads.
Coordinate occupancy and offset mask use categorical coding with transmitted CDFs.
Coordinate- and attribute-coder times are measured with host timers, without synchronization at every component boundary.
Context-network time is additionally measured with CUDA events.
These component times are diagnostic rather than an exclusive decomposition: a host interval may wait for GPU work already counted by the event timer, and asynchronously submitted work may finish outside that host interval.

\paragraph{Training time.}
The rate--distortion training timer starts after data and model initialization and stops after the optimization loop, before the final checkpoint save and evaluation.
It includes pruning, logging, periodic evaluation, and checkpoint writing.
The preceding 15K Scaffold-GS initialization is timed separately.

\subsection{Deterministic entropy decoding}
\label{app:cosa-integer-decoding}

\paragraph{Fixed-point representation.}
The integer variant uses a shared scale $S=2^{20}$: an int32 value $u$ represents the real value $u/S$.
This representation is used for context-network inputs and outputs, including reconstructed variables needed for subsequent predictions.
Linear layers use int8 weights and activations in $[-127,127]$, int32 biases and accumulators, and int64 intermediates for rescaling.
For integer $a$ and positive integer $d$, define
\begin{equation}
    \mathcal{R}(a,d)
    =\operatorname{sgn}(a)
      \left\lfloor\frac{|a|}{d}+\frac{1}{2}\right\rfloor .
    \label{eq:cosa-rounded-division}
\end{equation}
This denotes nearest rounding with half ties away from zero. This rounding is implemented using integer division, or a rounded right shift when the denominator is a power of two.
Parameter export and source-residual quantization instead use nearest rounding with half ties to even.

\paragraph{Coordinates and context networks.}
Octree decoding recovers origin-relative anchor coordinates in the same Morton order as the attribute streams.
For coordinate $c_{i,d}$ and transmitted coordinate range $L_d$, the normalized fixed-point input is
\begin{equation}
    u_{i,d}=\mathcal{R}(2S c_{i,d},L_d-1)-S,
    \qquad L_d>1.
    \label{eq:cosa-integer-coordinate}
\end{equation}
Geometry context is computed from these integers, without first reconstructing floating-point world coordinates.

As described in Sec.~\ref{sec:integer_entropy}, QAT uses per-tensor affine activation quantization and per-output-channel symmetric weight quantization.
Observers are frozen at iteration 25,000 in the default 30K configuration.
At export, the input zero-point correction is folded into the quantized bias, and scale conversions are represented by integer multipliers and shifts.
Hidden linear-layer outputs are converted to the shared fixed-point representation, passed through integer GELU, and requantized to int8 for the next layer.
Final network outputs remain in fixed point.
The GELU implementation tabulates $h(t)=t\Phi(-t)$ from Eq.~\eqref{eq:gelu_decomposition} over $[0,6]$ at spacing $1/512$, storing values with 24 fractional bits.
It uses integer linear interpolation and Eq.~\eqref{eq:cosa-rounded-division} to return to scale $S$.
The lookup table is shared by the encoder and decoder, so no floating-point evaluation of $\Phi$ is needed during entropy decoding.

\paragraph{Probability tables.}
The codec includes 128 fixed integer CDF tables for discretized zero-mean Gaussians, with standard deviations logarithmically spaced from $0.1$ to $256$.
They are generated offline and shared across scenes.
In the implementation, each distribution head predicts a mean and a continuous table index representing the standard deviation.
If the predicted index is stored as $v^{(q)}$ at scale $S$, the decoder selects
\begin{equation}
    \ell=\operatorname{clip}\!\left(\mathcal{R}(v^{(q)},S),0,127\right).
    \label{eq:cosa-integer-cdf-index}
\end{equation}
Thus, selecting a Gaussian distribution requires only integer rounding and clipping.
Coordinate and offset-mask CDFs are transmitted in the bitstream.

\paragraph{Causal reconstruction.}
For each group $g\in\{z,f,r,o,s\}$, export represents its quantization step by a multiplier $\rho_g$ and shift $\tau_g$ such that
\begin{equation}
    \Delta_g^{\mathrm{int}}=\frac{\rho_g}{S2^{\tau_g}}\approx\Delta_g.
    \label{eq:cosa-reconstruction-step}
\end{equation}
The encoder quantizes residuals using this effective step, and both sides use the transmitted pair $(\rho_g,\tau_g)$ for reconstruction.
For variables needed by later predictions, the fixed-point reconstruction is
\begin{equation}
    q_{\hat v}=q_\mu+\mathcal{R}(k_v \rho_g,2^{\tau_g}),
    \label{eq:cosa-causal-reconstruction}
\end{equation}
where $q_\mu$ is the fixed-point predicted mean and $k_v$ is the decoded residual.
This implements Eq.~\eqref{eq:integer_gaussian_reconstruction} without floating-point arithmetic.
In particular, previously reconstructed latent channels are used to predict the next channel, and reconstructed position scaling conditions the offset and Gaussian-scaling distributions.
Position and Gaussian scaling are coded in the logarithmic domain in the implementation.
All variables fed back into context prediction therefore remain in fixed point.

Features, offsets, and log-Gaussian scaling are not used for further entropy predictions.
Their final values are reconstructed in floating point as
\begin{equation}
    \hat v=\frac{q_\mu}{S}+k_v\Delta_g^{\mathrm{int}}.
    \label{eq:cosa-terminal-reconstruction}
\end{equation}
These final conversions do not affect entropy decoding.

\paragraph{Consistency scope.}
The scene bitstream stores the integer network parameters, rescaling parameters, reconstruction steps, coordinate metadata, mask CDF, compressed streams, and FP16 rendering-network parameters.
Decoding also requires the project source code, the model configuration files, and the shared Gaussian and GELU tables. These shared resources are not counted in the scene size.
For the same bitstream, consistency follows along the causal order: identical coordinates produce identical context inputs, the integer networks produce identical means and CDF indices, and the decoded residuals yield identical causal reconstructions.
Parallel entropy blocks write to fixed output slices, so their scheduling does not change the decoded sequence.
This guarantees consistency of entropy-decoded symbols and their fixed-point causal states, not bitwise equality of floating-point rendering.
In our experiments, we verify cross-platform consistency by checking exact equality of the entropy-decoded symbols for anchor latents, anchor attributes, offset masks, and the decoded integer anchor coordinates.

\subsection{Additional Rate--Distortion Results}
\label{sec:appendix_rd}

\begin{figure}[tb]
  \centering
  \includegraphics[width=0.328\linewidth,trim=5 6 5 5,clip]{fig/plots/mipnerf360_psnr.pdf}\hfill%
  \includegraphics[width=0.328\linewidth,trim=5 6 5 5,clip]{fig/plots/mipnerf360_ssim.pdf}\hfill%
  \includegraphics[width=0.328\linewidth,trim=5 6 5 5,clip]{fig/plots/mipnerf360_lpips.pdf}
  \\
  \includegraphics[width=0.328\linewidth,trim=5 6 5 5,clip]{fig/plots/deep_blending_psnr.pdf}\hfill%
  \includegraphics[width=0.328\linewidth,trim=5 6 5 5,clip]{fig/plots/deep_blending_ssim.pdf}\hfill%
  \includegraphics[width=0.328\linewidth,trim=5 6 5 5,clip]{fig/plots/deep_blending_lpips.pdf}
  \\
  \includegraphics[width=0.328\linewidth,trim=5 6 5 5,clip]{fig/plots/tandt_psnr.pdf}\hfill%
  \includegraphics[width=0.328\linewidth,trim=5 6 5 5,clip]{fig/plots/tandt_ssim.pdf}\hfill%
  \includegraphics[width=0.328\linewidth,trim=5 6 5 5,clip]{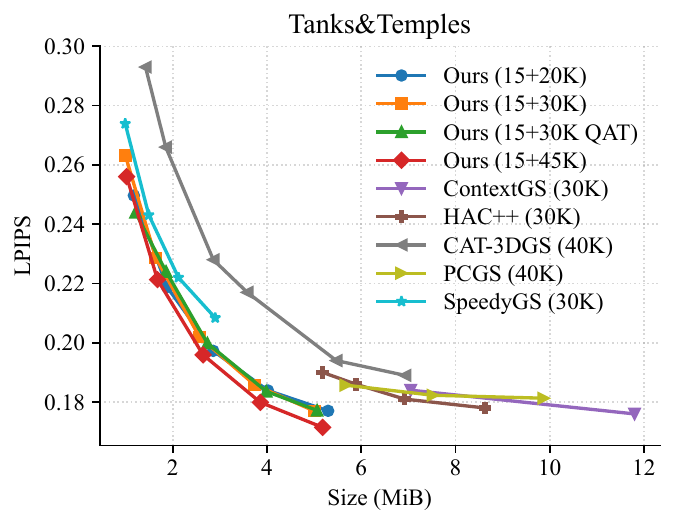}
  \\
  \includegraphics[width=0.328\linewidth,trim=5 6 5 5,clip]{fig/plots/bungeenerf_psnr.pdf}\hfill%
  \includegraphics[width=0.328\linewidth,trim=5 6 5 5,clip]{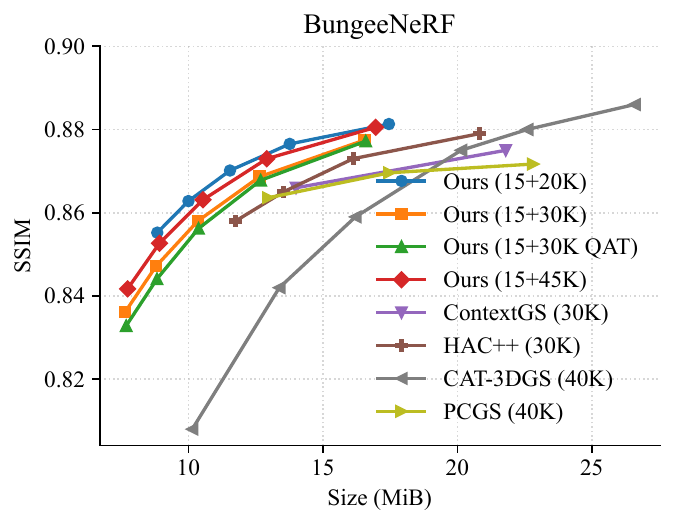}\hfill%
  \includegraphics[width=0.328\linewidth,trim=5 6 5 5,clip]{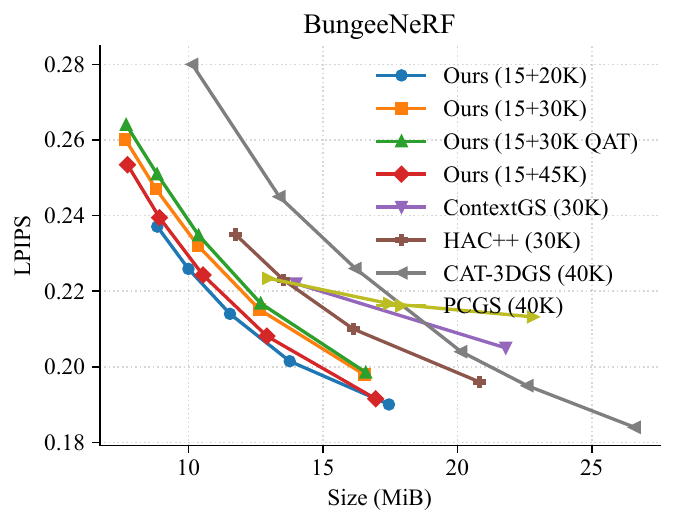}
  \caption{Rate--distortion comparison of \method{} and existing methods. Rows show Mip-NeRF360, DeepBlending, Tanks\&Temples, and BungeeNeRF, respectively. Columns show PSNR, SSIM, and LPIPS against scene size. }
  \label{Fig:RD_Appendix}
\end{figure}

Figure~\ref{Fig:RD_Appendix} presents PSNR, SSIM, and LPIPS against scene size on all four datasets.
Alongside existing methods, we compare the 15+45K, 15+30K, and 15+20K settings of \method{} and the 15+30K quantization-aware training (QAT) variant.

\paragraph{Optimization steps.}
Increasing the number of optimization steps generally improves the PSNR--size trade-off on Mip-NeRF360, DeepBlending, and Tanks\&Temples.
On Tanks\&Temples, the 15+45K setting also improves SSIM and LPIPS over much of the bitrate range.
On Mip-NeRF360 and DeepBlending, the SSIM--size and LPIPS--size curves remain close across the three training settings.
On BungeeNeRF, the 15+20K and 15+45K settings yield similar rate--distortion performance across all three metrics.

\paragraph{Quantization-aware training.}
The 15+30K QAT variant combines quantization-aware training with integer inference for the context model.
Its rate--distortion curves remain close to those of the floating-point 15+30K model across most bitrate points, with small quality losses at some low-bitrate points on DeepBlending and BungeeNeRF.

\begin{figure}[tb]
  \centering
  \includegraphics[width=0.328\linewidth,trim=5 6 5 5,clip]{fig/plots_ablation1/mipnerf360_psnr.pdf}\hfill%
  \includegraphics[width=0.328\linewidth,trim=5 6 5 5,clip]{fig/plots_ablation1/mipnerf360_ssim.pdf}\hfill%
  \includegraphics[width=0.328\linewidth,trim=5 6 5 5,clip]{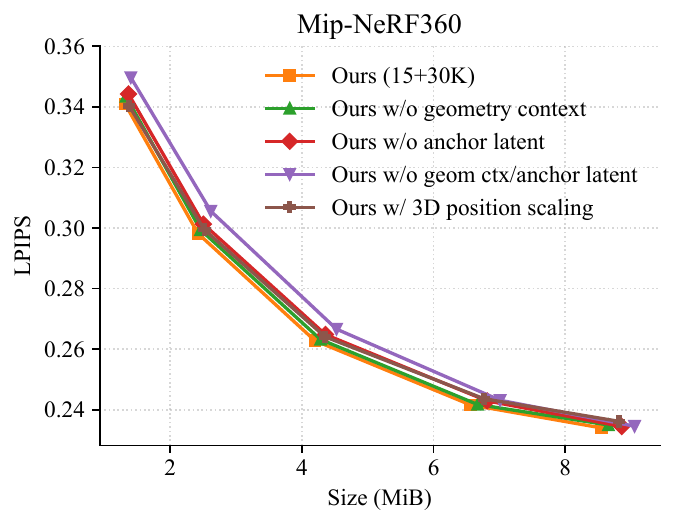}
  \\ \vspace{1.5em}
  \includegraphics[width=0.328\linewidth,trim=5 6 5 5,clip]{fig/plots_ablation2/mipnerf360_psnr.pdf}\hfill%
  \includegraphics[width=0.328\linewidth,trim=5 6 5 5,clip]{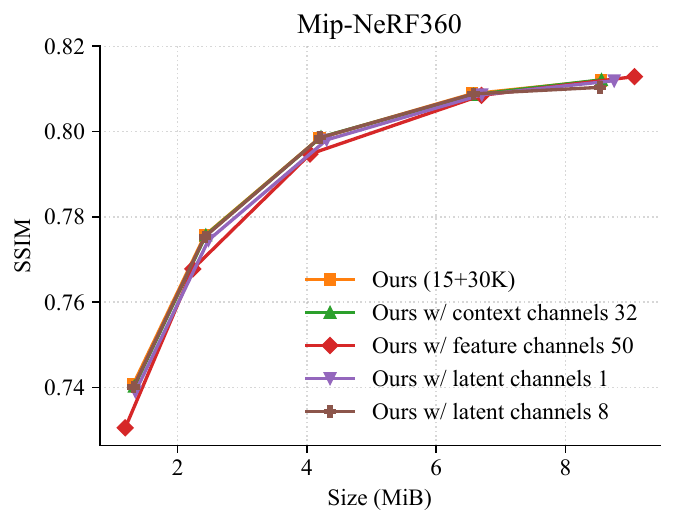}\hfill%
  \includegraphics[width=0.328\linewidth,trim=5 6 5 5,clip]{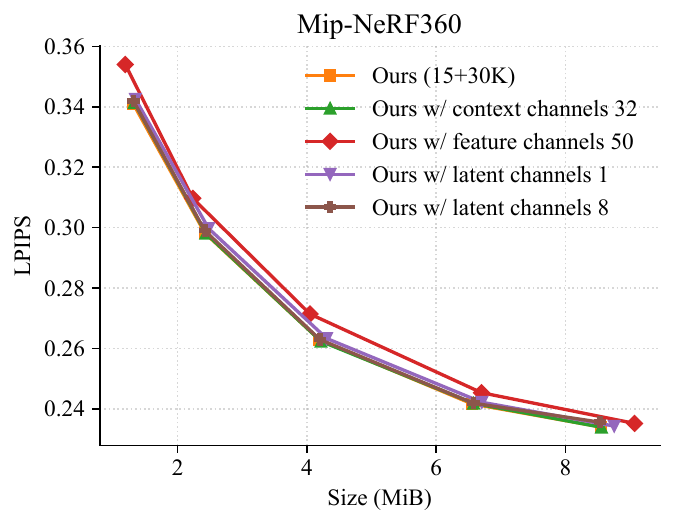}
  \vspace{0.5em}
\caption{Additional ablation results on Mip-NeRF360 under the 15+30K setting. Top: removing geometry context, anchor latents, or both, and replacing scalar position scaling with a three-dimensional vector. Bottom: varying the numbers of context, feature, and latent channels. Columns show PSNR, SSIM, and LPIPS against scene size.}
  \label{Fig:RD_Ablation_Appendix}
\end{figure}

\subsection{Additional Ablation Studies}
\label{sec:appendix_ablation}

Figure~\ref{Fig:RD_Ablation_Appendix} extends the main-paper ablations to all three quality metrics under the 15+30K setting.
The first row examines context sources and position scaling, while the second row examines the numbers of context, feature, and latent channels.

\paragraph{Context construction and position scaling.}
Removing both geometry context and anchor latents produces the clearest degradation at low bitrate points, with lower PSNR and SSIM and higher LPIPS.
Retaining either context source recovers much of this loss, showing that each provides informative context.
Combining the two sources yields a favorable overall rate--distortion trade-off.
Replacing scalar position scaling with a three-dimensional vector provides no consistent improvement across the three metrics.

\paragraph{Model capacity.}
Increasing the number of context channels from 24 to 32 has little effect on rate--distortion performance across the three metrics.
Increasing the number of feature channels from 32 to 50 slightly improves PSNR at the highest bitrate point but degrades performance across all three metrics at the lowest bitrate point.
Using one or eight latent channels yields rate--distortion curves close to those of the default four-channel setting.
These results show no consistent benefit from increasing the number of channels, supporting the default configuration.

\subsection{Detailed Results and Coding Times}
\label{sec:appendix_detailed_results}

Tables~\ref{tab:appendix_results_45k}, \ref{tab:appendix_results_30k}, \ref{tab:appendix_results_30k_qat}, and~\ref{tab:appendix_results_20k} report dataset-average reconstruction quality, compressed scene size, and total encoding and decoding times for the 15+45K, 15+30K, 15+30K QAT, and 15+20K settings, respectively.
Detailed per-scene results are provided in Tables~\ref{tab:appendix_scene_45k_mipnerf360}--\ref{tab:appendix_scene_45k_bungeenerf} for 15+45K, Tables~\ref{tab:appendix_scene_30k_mipnerf360}--\ref{tab:appendix_scene_30k_bungeenerf} for 15+30K, Tables~\ref{tab:appendix_scene_30k_qat_mipnerf360}--\ref{tab:appendix_scene_30k_qat_bungeenerf} for 15+30K QAT, and Tables~\ref{tab:appendix_scene_20k_mipnerf360}--\ref{tab:appendix_scene_20k_bungeenerf} for 15+20K.
Within each group, the tables report Mip-NeRF360, DeepBlending, Tanks\&Temples, and BungeeNeRF results in that order.
Runtime measurements use the reference platform with an NVIDIA 5880 Ada GPU and an AMD EPYC 9654 CPU. 
Table~\ref{tab:appendix_software_version} lists the software versions used on this platform. 

\paragraph{Detailed bitrate points.}
For each dataset, the 15+30K setting produces smaller bitstreams than the 15+20K setting at every tested rate weight $\lambda$.
Extending the number of rate--distortion optimization steps from 30K to 45K increases PSNR at every listed $\lambda$, with a small increase in scene size.
On BungeeNeRF, the 15+20K setting produces larger bitstreams and higher rendering quality than the 15+30K setting at the same $\lambda$, while the corresponding rate--distortion curves remain close.

\paragraph{Coding time.}
Across the three floating-point settings, dataset-average encoding times range from  0.0240 to 0.1752 s, and decoding times range from 0.0264 to 0.1839 s.
The dataset-average encoding and decoding times remain below 0.24 s at every tabulated bitrate point.

{
\begin{table}[tb]
  \centering
  \small
  \setlength{\tabcolsep}{5pt}
  \vspace{-1em}
  \caption{Dataset-average reconstruction quality, compressed scene size, and coding time of \method{} under the 15+45K setting. PSNR is measured in dB.}
  \vspace{1em}
  \label{tab:appendix_results_45k}
\begin{tabular}{crrrrrrr}
    \toprule
    Dataset & \multicolumn{1}{c}{$\lambda$} & \multicolumn{1}{c}{PSNR} & \multicolumn{1}{c}{SSIM} & \multicolumn{1}{c}{LPIPS} & \multicolumn{1}{c}{\shortstack{Size\\(MiB)}} & \multicolumn{1}{c}{\shortstack{Total Enc.\\(s)}} & \multicolumn{1}{c}{\shortstack{Total Dec.\\(s)}} \\
    \midrule
    \multirow{5}[2]{*}{Mip-NeRF360} & 0.0006 & 28.0353 & 0.8140 & 0.2299 & 8.7942 & 0.0925 & 0.1144 \\
        & 0.0010 & 27.9336 & 0.8109 & 0.2373 & 6.7579 & 0.0845 & 0.1061 \\
        & 0.0020 & 27.7153 & 0.8013 & 0.2567 & 4.3219 & 0.0652 & 0.0833 \\
        & 0.0040 & 27.2716 & 0.7796 & 0.2924 & 2.5008 & 0.0493 & 0.0657 \\
        & 0.0080 & 26.6699 & 0.7457 & 0.3348 & 1.3683 & 0.0375 & 0.0462 \\
    \midrule
    \multirow{5}[2]{*}{DeepBlending} & 0.0010 & 30.6693 & 0.9116 & 0.2514 & 3.4592 & 0.0451 & 0.0605 \\
        & 0.0020 & 30.6049 & 0.9110 & 0.2564 & 2.4439 & 0.0425 & 0.0525 \\
\bottomrule
    \end{tabular}%
\end{table}
}

{
\begin{table}[tb]
  \centering
  \small
  \setlength{\tabcolsep}{5pt}
  \vspace{-1em}
  \caption{Dataset-average reconstruction quality, compressed scene size, and coding time of \method{} under the 15+30K setting. PSNR is measured in dB.}
  \vspace{1em}
  \label{tab:appendix_results_30k}
    \begin{tabular}{crrrrrrr}
    \toprule
    Dataset & \multicolumn{1}{c}{$\lambda$} & \multicolumn{1}{c}{PSNR} & \multicolumn{1}{c}{SSIM} & \multicolumn{1}{c}{LPIPS} & \multicolumn{1}{c}{\shortstack{Size\\(MiB)}} & \multicolumn{1}{c}{\shortstack{Total Enc.\\(s)}} & \multicolumn{1}{c}{\shortstack{Total Dec.\\(s)}} \\
    \midrule
    \multirow{5}[2]{*}{Mip-NeRF360} & 0.0006 & 27.9403 & 0.8120 & 0.2340 & 8.5471 & 0.0895 & 0.1118 \\
        & 0.0010 & 27.8560 & 0.8089 & 0.2416 & 6.5575 & 0.0801 & 0.1012 \\
        & 0.0020 & 27.6126 & 0.7985 & 0.2628 & 4.1960 & 0.0636 & 0.0813 \\
        & 0.0040 & 27.1765 & 0.7757 & 0.2982 & 2.4203 & 0.0480 & 0.0619 \\
        & 0.0080 & 26.5370 & 0.7408 & 0.3409 & 1.3158 & 0.0356 & 0.0442 \\
    \midrule
    \multirow{5}[2]{*}{DeepBlending} & 0.0010 & 30.5756 & 0.9117 & 0.2525 & 3.3463 & 0.0448 & 0.0595 \\
        & 0.0020 & 30.5802 & 0.9101 & 0.2583 & 2.3720 & 0.0426 & 0.0553 \\
\bottomrule
    \end{tabular}%
\end{table}
}

{
\begin{table}[tb]
  \centering
  \small
  \setlength{\tabcolsep}{5pt}
  \vspace{-1em}
  \caption{Dataset-average reconstruction quality, compressed scene size, and coding time of \method{} under the 15+30K QAT setting. PSNR is measured in dB.}
  \label{tab:appendix_results_30k_qat}
  \vspace{1em}
    \begin{tabular}{crrrrrrr}
    \toprule
    Dataset & \multicolumn{1}{c}{$\lambda$} & \multicolumn{1}{c}{PSNR} & \multicolumn{1}{c}{SSIM} & \multicolumn{1}{c}{LPIPS} & \multicolumn{1}{c}{\shortstack{Size\\(MiB)}} & \multicolumn{1}{c}{\shortstack{Total Enc.\\(s)}} & \multicolumn{1}{c}{\shortstack{Total Dec.\\(s)}} \\
    \midrule
    \multirow{5}[2]{*}{Mip-NeRF360} & 0.0006 & 27.9519 & 0.8119 & 0.2344 & 8.6746 & 0.1068 & 0.1206 \\
        & 0.0010 & 27.8611 & 0.8089 & 0.2417 & 6.7016 & 0.1013 & 0.1099 \\
        & 0.0020 & 27.6498 & 0.8016 & 0.2579 & 4.4722 & 0.0894 & 0.0958 \\
        & 0.0040 & 27.2489 & 0.7822 & 0.2900 & 2.5831 & 0.0713 & 0.0762 \\
        & 0.0080 & 26.6939 & 0.7560 & 0.3249 & 1.4831 & 0.0609 & 0.0607 \\
    \midrule
    \multirow{5}[2]{*}{DeepBlending} & 0.0010 & 30.6162 & 0.9112 & 0.2532 & 3.4205 & 0.0624 & 0.0651 \\
        & 0.0020 & 30.5719 & 0.9104 & 0.2585 & 2.4474 & 0.0553 & 0.0555 \\
\bottomrule
    \end{tabular}%
\end{table}
}

{
\begin{table}[tb]
  \centering
  \small
  \setlength{\tabcolsep}{5pt}
  \vspace{-1em}
  \caption{Dataset-average reconstruction quality, compressed scene size, and coding time of \method{} under the 15+20K setting. PSNR is measured in dB.}
  \vspace{1em}
  \label{tab:appendix_results_20k}

    \begin{tabular}{crrrrrrr}
    \toprule
    Dataset & \multicolumn{1}{c}{$\lambda$} & \multicolumn{1}{c}{PSNR} & \multicolumn{1}{c}{SSIM} & \multicolumn{1}{c}{LPIPS} & \multicolumn{1}{c}{\shortstack{Size\\(MiB)}} & \multicolumn{1}{c}{\shortstack{Total Enc.\\(s)}} & \multicolumn{1}{c}{\shortstack{Total Dec.\\(s)}} \\
    \midrule
    \multirow{5}[2]{*}{Mip-NeRF360} & 0.0006 & 27.8221 & 0.8111 & 0.2333 & 9.1255 & 0.1024 & 0.1176 \\
        & 0.0010 & 27.7373 & 0.8088 & 0.2388 & 7.1409 & 0.0884 & 0.1131 \\
        & 0.0020 & 27.5516 & 0.8022 & 0.2530 & 4.7852 & 0.0728 & 0.0940 \\
        & 0.0040 & 27.2134 & 0.7870 & 0.2799 & 2.8861 & 0.0577 & 0.0717 \\
        & 0.0080 & 26.6940 & 0.7604 & 0.3167 & 1.6292 & 0.0435 & 0.0542 \\
    \midrule
    \multirow{5}[2]{*}{DeepBlending} & 0.0010 & 30.5821 & 0.9114 & 0.2535 & 3.5288 & 0.0461 & 0.0616 \\
        & 0.0020 & 30.4643 & 0.9101 & 0.2585 & 2.5626 & 0.0423 & 0.0550 \\
\bottomrule
    \end{tabular}%
\end{table}
}

{
\begin{table}[htbp]
  \centering
  \small
  \setlength{\tabcolsep}{6pt}
  \vspace{-1em}
  \caption{Scene-wise reconstruction quality, compressed scene size, and coding time of \method{} under the 15+45K setting on Mip-NeRF360. PSNR is measured in dB.}
  \label{tab:appendix_scene_45k_mipnerf360}
  \vspace{1em}
    \begin{tabular}{lcrrrrrr}
    \toprule
    Scene & $\lambda$ & PSNR & SSIM & LPIPS & Size (MiB) & \shortstack{Total Enc.\\(s)} & \shortstack{Total Dec.\\(s)} \\
    \midrule
    bicycle & \multirow{9}[2]{*}{0.0006} & 25.2835 & 0.7478 & 0.2630 & 12.0663 & 0.1225 & 0.1521 \\
    bonsai &     & 33.2550 & 0.9499 & 0.1793 & 5.3596 & 0.0730 & 0.0851 \\
    counter &     & 30.1227 & 0.9201 & 0.1821 & 5.9570 & 0.0600 & 0.0804 \\
    flowers &     & 21.4197 & 0.5845 & 0.3740 & 10.2681 & 0.1022 & 0.1288 \\
    garden &     & 27.4425 & 0.8475 & 0.1485 & 12.0146 & 0.1177 & 0.1481 \\
    kitchen &     & 32.1876 & 0.9330 & 0.1196 & 6.8020 & 0.0732 & 0.0881 \\
    room &     & 32.6127 & 0.9315 & 0.1897 & 4.5917 & 0.0527 & 0.0732 \\
\bottomrule
    \end{tabular}%
\end{table}%
}

{
\begin{table}[htbp]
  \centering
  \small
  \setlength{\tabcolsep}{6pt}
  \vspace{-1em}
  \caption{Scene-wise reconstruction quality, compressed scene size, and coding time of \method{} under the 15+45K setting on DeepBlending. PSNR is measured in dB.}
  \label{tab:appendix_scene_45k_deep_blending}
  \vspace{1em}
    \begin{tabular}{lcrrrrrr}
    \toprule
    Scene & $\lambda$ & PSNR & SSIM & LPIPS & Size (MiB) & \shortstack{Total Enc.\\(s)} & \shortstack{Total Dec.\\(s)} \\
    \midrule
    drjohnson & \multirow{2}[2]{*}{0.0010} & 30.0368 & 0.9088 & 0.2509 & 3.8020 & 0.0475 & 0.0641 \\
    playroom &     & 31.3019 & 0.9144 & 0.2519 & 3.1163 & 0.0427 & 0.0568 \\
    \midrule
    drjohnson & \multirow{2}[2]{*}{0.0020} & 30.0120 & 0.9087 & 0.2544 & 2.6861 & 0.0489 & 0.0579 \\
    playroom &     & 31.1977 & 0.9133 & 0.2584 & 2.2017 & 0.0362 & 0.0471 \\
    \midrule
    drjohnson & \multirow{2}[2]{*}{0.0040} & 29.8398 & 0.9043 & 0.2659 & 1.7732 & 0.0394 & 0.0514 \\
    playroom &     & 31.1085 & 0.9109 & 0.2671 & 1.4331 & 0.0395 & 0.0395 \\
    \midrule
    drjohnson & \multirow{2}[2]{*}{0.0080} & 29.4591 & 0.8959 & 0.2846 & 1.1316 & 0.0309 & 0.0378 \\
    playroom &     & 30.7960 & 0.9064 & 0.2808 & 0.9152 & 0.0282 & 0.0314 \\
    \midrule
    drjohnson & \multirow{2}[2]{*}{0.0160} & 28.9196 & 0.8841 & 0.3059 & 0.7131 & 0.0257 & 0.0284 \\
    playroom &     & 30.3566 & 0.8989 & 0.3002 & 0.5814 & 0.0224 & 0.0244 \\
    \bottomrule
    \end{tabular}%
\end{table}%
}

{
\begin{table}[htbp]
  \centering
  \small
  \setlength{\tabcolsep}{6pt}
  \vspace{-1em}
  \caption{Scene-wise reconstruction quality, compressed scene size, and coding time of \method{} under the 15+45K setting on Tanks\&Temples. PSNR is measured in dB.}
  \label{tab:appendix_scene_45k_tandt}
  \vspace{1em}
    \begin{tabular}{lcrrrrrr}
    \toprule
    Scene & $\lambda$ & PSNR & SSIM & LPIPS & Size (MiB) & \shortstack{Total Enc.\\(s)} & \shortstack{Total Dec.\\(s)} \\
    \midrule
    train & \multirow{2}[2]{*}{0.0010} & 22.9447 & 0.8288 & 0.2014 & 5.2069 & 0.0588 & 0.0759 \\
    truck &     & 26.2113 & 0.8893 & 0.1415 & 5.1546 & 0.0618 & 0.0823 \\
    \midrule
    train & \multirow{2}[2]{*}{0.0020} & 22.7386 & 0.8246 & 0.2082 & 4.0864 & 0.0472 & 0.0708 \\
    truck &     & 26.1053 & 0.8854 & 0.1516 & 3.6329 & 0.0525 & 0.0737 \\
    \midrule
    train & \multirow{2}[2]{*}{0.0040} & 22.7198 & 0.8194 & 0.2206 & 2.9058 & 0.0443 & 0.0590 \\
    truck &     & 25.8799 & 0.8771 & 0.1713 & 2.3787 & 0.0455 & 0.0558 \\
    \midrule
    train & \multirow{2}[2]{*}{0.0080} & 22.5913 & 0.8067 & 0.2421 & 1.8879 & 0.0376 & 0.0469 \\
    truck &     & 25.5414 & 0.8626 & 0.2005 & 1.4621 & 0.0329 & 0.0467 \\
    \midrule
    train & \multirow{2}[2]{*}{0.0160} & 22.1488 & 0.7839 & 0.2745 & 1.1577 & 0.0314 & 0.0439 \\
    truck &     & 24.9598 & 0.8415 & 0.2377 & 0.8828 & 0.0311 & 0.0348 \\
    \bottomrule
    \end{tabular}%
\end{table}%
}

{
\begin{table}[htbp]
  \centering
  \small
  \setlength{\tabcolsep}{6pt}
  \vspace{-1em}
  \caption{Scene-wise reconstruction quality, compressed scene size, and coding time of \method{} under the 15+45K setting on BungeeNeRF. PSNR is measured in dB.}
  \label{tab:appendix_scene_45k_bungeenerf}
  \vspace{1em}
    \begin{tabular}{lcrrrrrr}
    \toprule
    Scene & $\lambda$ & PSNR & SSIM & LPIPS & Size (MiB) & \shortstack{Total Enc.\\(s)} & \shortstack{Total Dec.\\(s)} \\
    \midrule
    amsterdam & \multirow{6}[2]{*}{0.0002} & 27.4387 & 0.9010 & 0.1643 & 18.2302 & 0.1678 & 0.2007 \\
    bilbao &     & 27.8780 & 0.8871 & 0.1782 & 16.9360 & 0.1592 & 0.1693 \\
    hollywood &     & 24.8905 & 0.7974 & 0.2851 & 17.0617 & 0.1647 & 0.1800 \\
    pompidou &     & 25.8096 & 0.8631 & 0.2114 & 17.6858 & 0.1620 & 0.1643 \\
    quebec &     & 30.6954 & 0.9450 & 0.1333 & 14.3947 & 0.1385 & 0.1504 \\
    rome &     & 26.9121 & 0.8895 & 0.1770 & 17.4639 & 0.1832 & 0.2005 \\
    \midrule
    amsterdam & \multirow{6}[2]{*}{0.0004} & 27.2421 & 0.8889 & 0.1899 & 13.5120 & 0.1410 & 0.1621 \\
\bottomrule
    \end{tabular}%
\end{table}%
}

{
\begin{table}[htbp]
  \centering
  \small
  \setlength{\tabcolsep}{6pt}
  \vspace{-1em}
  \caption{Scene-wise reconstruction quality, compressed scene size, and coding time of \method{} under the 15+30K setting on Mip-NeRF360. PSNR is measured in dB.}
  \label{tab:appendix_scene_30k_mipnerf360}
  \vspace{1em}
    \begin{tabular}{lcrrrrrr}
    \toprule
    Scene & $\lambda$ & PSNR & SSIM & LPIPS & Size (MiB) & \shortstack{Total Enc.\\(s)} & \shortstack{Total Dec.\\(s)} \\
    \midrule
    bicycle & \multirow{9}[2]{*}{0.0006} & 25.1977 & 0.7457 & 0.2677 & 11.6525 & 0.1280 & 0.1498 \\
    bonsai &     & 33.1701 & 0.9493 & 0.1813 & 5.2926 & 0.0691 & 0.0865 \\
    counter &     & 29.9499 & 0.9185 & 0.1851 & 5.8135 & 0.0579 & 0.0781 \\
    flowers &     & 21.4071 & 0.5816 & 0.3784 & 9.9852 & 0.1009 & 0.1279 \\
    garden &     & 27.3482 & 0.8442 & 0.1551 & 11.6989 & 0.1169 & 0.1352 \\
    kitchen &     & 32.0032 & 0.9306 & 0.1225 & 6.6229 & 0.0757 & 0.0878 \\
    room &     & 32.4597 & 0.9301 & 0.1934 & 4.4935 & 0.0525 & 0.0717 \\
\bottomrule
    \end{tabular}%
\end{table}%
}

{
\begin{table}[htbp]
  \centering
  \small
  \setlength{\tabcolsep}{6pt}
  \vspace{-1em}
  \caption{Scene-wise reconstruction quality, compressed scene size, and coding time of \method{} under the 15+30K setting on DeepBlending. PSNR is measured in dB.}
  \label{tab:appendix_scene_30k_deep_blending}
  \vspace{1em}
    \begin{tabular}{lcrrrrrr}
    \toprule
    Scene & $\lambda$ & PSNR & SSIM & LPIPS & Size (MiB) & \shortstack{Total Enc.\\(s)} & \shortstack{Total Dec.\\(s)} \\
    \midrule
    drjohnson & \multirow{2}[2]{*}{0.0010} & 29.9732 & 0.9093 & 0.2519 & 3.6608 & 0.0482 & 0.0638 \\
    playroom &     & 31.1780 & 0.9142 & 0.2532 & 3.0318 & 0.0414 & 0.0553 \\
    \midrule
    drjohnson & \multirow{2}[2]{*}{0.0020} & 29.9998 & 0.9077 & 0.2574 & 2.6012 & 0.0463 & 0.0545 \\
    playroom &     & 31.1607 & 0.9125 & 0.2591 & 2.1429 & 0.0389 & 0.0561 \\
    \midrule
    drjohnson & \multirow{2}[2]{*}{0.0040} & 29.8081 & 0.9040 & 0.2671 & 1.7314 & 0.0347 & 0.0444 \\
    playroom &     & 30.9881 & 0.9094 & 0.2708 & 1.4326 & 0.0327 & 0.0378 \\
    \midrule
    drjohnson & \multirow{2}[2]{*}{0.0080} & 29.4251 & 0.8962 & 0.2856 & 1.1180 & 0.0303 & 0.0351 \\
    playroom &     & 30.7169 & 0.9054 & 0.2827 & 0.9112 & 0.0257 & 0.0301 \\
    \midrule
    drjohnson & \multirow{2}[2]{*}{0.0160} & 28.9561 & 0.8864 & 0.3045 & 0.6970 & 0.0262 & 0.0343 \\
    playroom &     & 30.2921 & 0.8982 & 0.3005 & 0.5795 & 0.0224 & 0.0272 \\
    \bottomrule
    \end{tabular}%
\end{table}%
}

{
\begin{table}[htbp]
  \centering
  \small
  \setlength{\tabcolsep}{6pt}
  \vspace{-1em}
  \caption{Scene-wise reconstruction quality, compressed scene size, and coding time of \method{} under the 15+30K setting on Tanks\&Temples. PSNR is measured in dB.}
  \label{tab:appendix_scene_30k_tandt}
  \vspace{1em}
    \begin{tabular}{lcrrrrrr}
    \toprule
    Scene & $\lambda$ & PSNR & SSIM & LPIPS & Size (MiB) & \shortstack{Total Enc.\\(s)} & \shortstack{Total Dec.\\(s)} \\
    \midrule
    train & \multirow{2}[2]{*}{0.0010} & 22.8962 & 0.8260 & 0.2069 & 4.9931 & 0.0581 & 0.0723 \\
    truck &     & 26.0951 & 0.8871 & 0.1469 & 5.0286 & 0.0582 & 0.0907 \\
    \midrule
    train & \multirow{2}[2]{*}{0.0020} & 22.6636 & 0.8229 & 0.2137 & 3.9408 & 0.0481 & 0.0696 \\
    truck &     & 25.9664 & 0.8827 & 0.1580 & 3.5281 & 0.0542 & 0.0701 \\
    \midrule
    train & \multirow{2}[2]{*}{0.0040} & 22.7186 & 0.8160 & 0.2267 & 2.8295 & 0.0438 & 0.0569 \\
    truck &     & 25.7244 & 0.8736 & 0.1770 & 2.3080 & 0.0436 & 0.0552 \\
    \midrule
    train & \multirow{2}[2]{*}{0.0080} & 22.3242 & 0.8011 & 0.2500 & 1.8518 & 0.0353 & 0.0450 \\
    truck &     & 25.3941 & 0.8589 & 0.2074 & 1.4238 & 0.0346 & 0.0426 \\
    \midrule
    train & \multirow{2}[2]{*}{0.0160} & 21.9496 & 0.7782 & 0.2809 & 1.1275 & 0.0296 & 0.0348 \\
    truck &     & 24.8685 & 0.8363 & 0.2454 & 0.8591 & 0.0313 & 0.0313 \\
    \bottomrule
    \end{tabular}%
\end{table}%
}

{
\begin{table}[htbp]
  \centering
  \small
  \setlength{\tabcolsep}{6pt}
  \vspace{-1em}
  \caption{Scene-wise reconstruction quality, compressed scene size, and coding time of \method{} under the 15+30K setting on BungeeNeRF. PSNR is measured in dB.}
  \label{tab:appendix_scene_30k_bungeenerf}
  \vspace{1em}
    \begin{tabular}{lcrrrrrr}
    \toprule
    Scene & $\lambda$ & PSNR & SSIM & LPIPS & Size (MiB) & \shortstack{Total Enc.\\(s)} & \shortstack{Total Dec.\\(s)} \\
    \midrule
    amsterdam & \multirow{6}[2]{*}{0.0002} & 27.2770 & 0.8953 & 0.1749 & 17.7631 & 0.1830 & 0.1868 \\
    bilbao &     & 27.8415 & 0.8852 & 0.1827 & 16.5058 & 0.1532 & 0.1831 \\
    hollywood &     & 24.8069 & 0.7917 & 0.2969 & 16.5321 & 0.1488 & 0.1635 \\
    pompidou &     & 25.6741 & 0.8595 & 0.2166 & 17.2445 & 0.1768 & 0.1688 \\
    quebec &     & 30.7004 & 0.9445 & 0.1355 & 14.1288 & 0.1334 & 0.1486 \\
    rome &     & 26.9417 & 0.8884 & 0.1809 & 17.0978 & 0.1628 & 0.1975 \\
    \midrule
    amsterdam & \multirow{6}[2]{*}{0.0004} & 27.0565 & 0.8819 & 0.2014 & 13.2343 & 0.1689 & 0.1572 \\
\bottomrule
    \end{tabular}%
\end{table}%
}

{
\begin{table}[htbp]
  \centering
  \small
  \setlength{\tabcolsep}{6pt}
  \vspace{-1em}
  \caption{Scene-wise reconstruction quality, compressed scene size, and coding time of \method{} under the 15+30K QAT setting on Mip-NeRF360. PSNR is measured in dB.}
  \label{tab:appendix_scene_30k_qat_mipnerf360}
  \vspace{1em}

    \begin{tabular}{lcrrrrrr}
    \toprule
    Scene & $\lambda$ & PSNR & SSIM & LPIPS & Size (MiB) & \shortstack{Total Enc.\\(s)} & \shortstack{Total Dec.\\(s)} \\
    \midrule
    bicycle & \multirow{9}[2]{*}{0.0006} & 25.2507 & 0.7457 & 0.2681 & 11.9511 & 0.1469 & 0.1661 \\
    bonsai &     & 33.1603 & 0.9491 & 0.1814 & 5.4509 & 0.0845 & 0.0908 \\
    counter &     & 29.9500 & 0.9183 & 0.1854 & 5.8636 & 0.0770 & 0.0864 \\
    flowers &     & 21.3977 & 0.5812 & 0.3790 & 10.0742 & 0.1139 & 0.1361 \\
    garden &     & 27.3605 & 0.8438 & 0.1561 & 11.7687 & 0.1410 & 0.1487 \\
    kitchen &     & 31.9857 & 0.9307 & 0.1227 & 6.7252 & 0.0857 & 0.0986 \\
    room &     & 32.4843 & 0.9303 & 0.1936 & 4.5560 & 0.0682 & 0.0780 \\
\bottomrule
    \end{tabular}%
\end{table}%
}

{
\begin{table}[htbp]
  \centering
  \small
  \setlength{\tabcolsep}{6pt}
  \vspace{-1em}
  \caption{Scene-wise reconstruction quality, compressed scene size, and coding time of \method{} under the 15+30K QAT setting on DeepBlending. PSNR is measured in dB.}
  \label{tab:appendix_scene_30k_qat_deep_blending}
  \vspace{1em}
    \begin{tabular}{lcrrrrrr}
    \toprule
    Scene & $\lambda$ & PSNR & SSIM & LPIPS & Size (MiB) & \shortstack{Total Enc.\\(s)} & \shortstack{Total Dec.\\(s)} \\
    \midrule
    drjohnson & \multirow{2}[2]{*}{0.0010} & 29.9794 & 0.9083 & 0.2535 & 3.7563 & 0.0651 & 0.0693 \\
    playroom &     & 31.2529 & 0.9140 & 0.2530 & 3.0848 & 0.0597 & 0.0610 \\
    \midrule
    drjohnson & \multirow{2}[2]{*}{0.0020} & 29.9069 & 0.9077 & 0.2580 & 2.6702 & 0.0576 & 0.0592 \\
    playroom &     & 31.2368 & 0.9131 & 0.2591 & 2.2246 & 0.0529 & 0.0518 \\
    \midrule
    drjohnson & \multirow{2}[2]{*}{0.0040} & 29.7257 & 0.9040 & 0.2668 & 1.8523 & 0.0536 & 0.0522 \\
    playroom &     & 30.9665 & 0.9099 & 0.2694 & 1.4402 & 0.0502 & 0.0431 \\
    \midrule
    drjohnson & \multirow{2}[2]{*}{0.0080} & 29.4492 & 0.8966 & 0.2849 & 1.1217 & 0.0460 & 0.0401 \\
    playroom &     & 30.6247 & 0.9045 & 0.2838 & 0.9539 & 0.0454 & 0.0390 \\
    \midrule
    drjohnson & \multirow{2}[2]{*}{0.0160} & 28.7005 & 0.8844 & 0.3080 & 0.7685 & 0.0434 & 0.0380 \\
    playroom &     & 30.1098 & 0.8974 & 0.3013 & 0.6472 & 0.0415 & 0.0330 \\
    \bottomrule
    \end{tabular}%
\end{table}%
}

{
\begin{table}[htbp]
  \centering
  \small
  \setlength{\tabcolsep}{6pt}
  \vspace{-1em}
  \caption{Scene-wise reconstruction quality, compressed scene size, and coding time of \method{} under the 15+30K QAT setting on Tanks\&Temples. PSNR is measured in dB.}
  \label{tab:appendix_scene_30k_qat_tandt}
  \vspace{1em}
    \begin{tabular}{lcrrrrrr}
    \toprule
    Scene & $\lambda$ & PSNR & SSIM & LPIPS & Size (MiB) & \shortstack{Total Enc.\\(s)} & \shortstack{Total Dec.\\(s)} \\
    \midrule
    train & \multirow{2}[2]{*}{0.0010} & 22.8444 & 0.8267 & 0.2069 & 5.0423 & 0.0754 & 0.0799 \\
    truck &     & 26.0748 & 0.8867 & 0.1475 & 5.0854 & 0.0873 & 0.0882 \\
    \midrule
    train & \multirow{2}[2]{*}{0.0020} & 22.7016 & 0.8231 & 0.2138 & 4.0165 & 0.0739 & 0.0716 \\
    truck &     & 26.0107 & 0.8840 & 0.1534 & 3.9499 & 0.0931 & 0.0932 \\
    \midrule
    train & \multirow{2}[2]{*}{0.0040} & 22.4999 & 0.8148 & 0.2286 & 2.8787 & 0.0633 & 0.0615 \\
    truck &     & 25.7366 & 0.8757 & 0.1712 & 2.5863 & 0.0754 & 0.0698 \\
    \midrule
    train & \multirow{2}[2]{*}{0.0080} & 22.2251 & 0.7986 & 0.2544 & 2.0732 & 0.0601 & 0.0502 \\
    truck &     & 25.4420 & 0.8644 & 0.1936 & 1.6395 & 0.0636 & 0.0589 \\
    \midrule
    train & \multirow{2}[2]{*}{0.0160} & 21.9164 & 0.7840 & 0.2691 & 1.2876 & 0.0559 & 0.0467 \\
    truck &     & 24.8872 & 0.8492 & 0.2189 & 1.1300 & 0.0616 & 0.0531 \\
    \bottomrule
    \end{tabular}%
\end{table}%
}

{
\begin{table}[htbp]
  \centering
  \small
  \setlength{\tabcolsep}{6pt}
  \vspace{-1em}
  \caption{Scene-wise reconstruction quality, compressed scene size, and coding time of \method{} under the 15+30K QAT setting on BungeeNeRF. PSNR is measured in dB.}
  \label{tab:appendix_scene_30k_qat_bungeenerf}
  \vspace{1em}
    \begin{tabular}{lcrrrrrr}
    \toprule
    Scene & $\lambda$ & PSNR & SSIM & LPIPS & Size (MiB) & \shortstack{Total Enc.\\(s)} & \shortstack{Total Dec.\\(s)} \\
    \midrule
    amsterdam & \multirow{6}[2]{*}{0.0002} & 27.2753 & 0.8947 & 0.1763 & 17.8597 & 0.2041 & 0.2331 \\
    bilbao &     & 27.8726 & 0.8854 & 0.1836 & 16.5345 & 0.1899 & 0.2020 \\
    hollywood &     & 24.7648 & 0.7916 & 0.2983 & 16.5427 & 0.1871 & 0.1848 \\
    pompidou &     & 25.6674 & 0.8592 & 0.2163 & 17.2884 & 0.2035 & 0.2068 \\
    quebec &     & 30.6520 & 0.9441 & 0.1357 & 14.1433 & 0.1546 & 0.1810 \\
    rome &     & 26.9374 & 0.8882 & 0.1809 & 17.1984 & 0.1861 & 0.1967 \\
    \midrule
    amsterdam & \multirow{6}[2]{*}{0.0004} & 27.0388 & 0.8802 & 0.2038 & 13.2845 & 0.1747 & 0.1726 \\
\bottomrule
    \end{tabular}%
\end{table}%
}

{
\begin{table}[htbp]
  \centering
  \small
  \setlength{\tabcolsep}{6pt}
  \vspace{-1em}
  \caption{Scene-wise reconstruction quality, compressed scene size, and coding time of \method{} under the 15+20K setting on Mip-NeRF360. PSNR is measured in dB.}
  \label{tab:appendix_scene_20k_mipnerf360}
  \vspace{1em}
    \begin{tabular}{lcrrrrrr}
    \toprule
    Scene & $\lambda$ & PSNR & SSIM & LPIPS & Size (MiB) & \shortstack{Total Enc.\\(s)} & \shortstack{Total Dec.\\(s)} \\
    \midrule
    bicycle & \multirow{9}[2]{*}{0.0006} & 25.0466 & 0.7421 & 0.2671 & 12.5481 & 0.1484 & 0.1565 \\
    bonsai &     & 32.7906 & 0.9477 & 0.1823 & 5.6182 & 0.0715 & 0.0896 \\
    counter &     & 29.7514 & 0.9173 & 0.1867 & 6.0469 & 0.0613 & 0.0812 \\
    flowers &     & 21.3726 & 0.5797 & 0.3773 & 10.9682 & 0.1229 & 0.1469 \\
    garden &     & 27.4057 & 0.8461 & 0.1495 & 12.7902 & 0.1296 & 0.1450 \\
    kitchen &     & 31.8157 & 0.9295 & 0.1246 & 6.9196 & 0.0906 & 0.0904 \\
    room &     & 32.2352 & 0.9297 & 0.1925 & 4.6814 & 0.0567 & 0.0747 \\
\bottomrule
    \end{tabular}%
\end{table}%
}

{
\begin{table}[htbp]
  \centering
  \small
  \setlength{\tabcolsep}{6pt}
  \vspace{-1em}
  \caption{Scene-wise reconstruction quality, compressed scene size, and coding time of \method{} under the 15+20K setting on DeepBlending. PSNR is measured in dB.}
  \label{tab:appendix_scene_20k_deep_blending}
  \vspace{1em}
     \begin{tabular}{lcrrrrrr}
    \toprule
    Scene & $\lambda$ & PSNR & SSIM & LPIPS & Size (MiB) & \shortstack{Total Enc.\\(s)} & \shortstack{Total Dec.\\(s)} \\
    \midrule
    drjohnson & \multirow{2}[2]{*}{0.0010} & 29.9249 & 0.9089 & 0.2528 & 3.8178 & 0.0479 & 0.0648 \\
    playroom &     & 31.2393 & 0.9139 & 0.2543 & 3.2398 & 0.0442 & 0.0584 \\
    \midrule
    drjohnson & \multirow{2}[2]{*}{0.0020} & 29.8601 & 0.9074 & 0.2576 & 2.7912 & 0.0450 & 0.0593 \\
    playroom &     & 31.0684 & 0.9128 & 0.2595 & 2.3341 & 0.0396 & 0.0507 \\
    \midrule
    drjohnson & \multirow{2}[2]{*}{0.0040} & 29.6666 & 0.9036 & 0.2674 & 1.9119 & 0.0372 & 0.0486 \\
    playroom &     & 30.9235 & 0.9107 & 0.2676 & 1.5874 & 0.0341 & 0.0429 \\
    \midrule
    drjohnson & \multirow{2}[2]{*}{0.0080} & 29.3999 & 0.8984 & 0.2795 & 1.2449 & 0.0332 & 0.0411 \\
    playroom &     & 30.7157 & 0.9065 & 0.2799 & 1.0398 & 0.0353 & 0.0354 \\
    \midrule
    drjohnson & \multirow{2}[2]{*}{0.0160} & 28.9911 & 0.8897 & 0.2988 & 0.8064 & 0.0280 & 0.0327 \\
    playroom &     & 30.1847 & 0.8994 & 0.2965 & 0.6782 & 0.0248 & 0.0290 \\
    \bottomrule
    \end{tabular}%
\end{table}%
}

{
\begin{table}[htbp]
  \centering
  \small
  \setlength{\tabcolsep}{6pt}
  \vspace{-1em}
  \caption{Scene-wise reconstruction quality, compressed scene size, and coding time of \method{} under the 15+20K setting on Tanks\&Temples. PSNR is measured in dB.}
  \label{tab:appendix_scene_20k_tandt}
  \vspace{1em}
    \begin{tabular}{lcrrrrrr}
    \toprule
    Scene & $\lambda$ & PSNR & SSIM & LPIPS & Size (MiB) & \shortstack{Total Enc.\\(s)} & \shortstack{Total Dec.\\(s)} \\
    \midrule
    train & \multirow{2}[2]{*}{0.0010} & 22.9435 & 0.8250 & 0.2083 & 5.2135 & 0.0646 & 0.0766 \\
    truck &     & 26.0186 & 0.8863 & 0.1458 & 5.3790 & 0.0751 & 0.0876 \\
    \midrule
    train & \multirow{2}[2]{*}{0.0020} & 22.6356 & 0.8207 & 0.2141 & 4.1309 & 0.0490 & 0.0724 \\
    truck &     & 25.9549 & 0.8834 & 0.1537 & 3.9056 & 0.0572 & 0.0823 \\
    \midrule
    train & \multirow{2}[2]{*}{0.0040} & 22.3971 & 0.8142 & 0.2250 & 3.1066 & 0.0474 & 0.0586 \\
    truck &     & 25.7069 & 0.8756 & 0.1695 & 2.6080 & 0.0521 & 0.0634 \\
    \midrule
    train & \multirow{2}[2]{*}{0.0080} & 22.2610 & 0.8029 & 0.2443 & 2.0930 & 0.0400 & 0.0537 \\
    truck &     & 25.3989 & 0.8642 & 0.1932 & 1.6638 & 0.0394 & 0.0495 \\
    \midrule
    train & \multirow{2}[2]{*}{0.0160} & 21.8626 & 0.7827 & 0.2729 & 1.3114 & 0.0328 & 0.0384 \\
    truck &     & 24.9076 & 0.8456 & 0.2265 & 1.0359 & 0.0305 & 0.0379 \\
    \bottomrule
    \end{tabular}%
\end{table}%
}

{
\begin{table}[htbp]
  \centering
  \small
  \setlength{\tabcolsep}{6pt}
  \vspace{-1em}
  \caption{Scene-wise reconstruction quality, compressed scene size, and coding time of \method{} under the 15+20K setting on BungeeNeRF. PSNR is measured in dB.}
  \label{tab:appendix_scene_20k_bungeenerf}
  \vspace{1em}
    \begin{tabular}{lcrrrrrr}
    \toprule
    Scene & $\lambda$ & PSNR & SSIM & LPIPS & Size (MiB) & \shortstack{Total Enc.\\(s)} & \shortstack{Total Dec.\\(s)} \\
    \midrule
    amsterdam & \multirow{6}[2]{*}{0.0002} & 27.4180 & 0.9009 & 0.1634 & 18.9424 & 0.1912 & 0.2115 \\
    bilbao &     & 27.9942 & 0.8909 & 0.1706 & 17.4101 & 0.1664 & 0.1690 \\
    hollywood &     & 24.8852 & 0.7978 & 0.2853 & 17.2021 & 0.1672 & 0.1867 \\
    pompidou &     & 25.8569 & 0.8643 & 0.2100 & 18.3461 & 0.1986 & 0.1788 \\
    quebec &     & 30.6481 & 0.9445 & 0.1343 & 14.8068 & 0.1484 & 0.1563 \\
    rome &     & 26.8999 & 0.8894 & 0.1765 & 18.0249 & 0.1794 & 0.2009 \\
    \midrule
    amsterdam & \multirow{6}[2]{*}{0.0004} & 27.2869 & 0.8930 & 0.1811 & 14.6458 & 0.1692 & 0.1794 \\
\bottomrule
    \end{tabular}%
\end{table}%
}

\begin{table}[t]
    \centering
    \small
      \setlength{\tabcolsep}{22pt}
    \caption{Software versions of the default platform with an NVIDIA 5880 Ada GPU and an AMD EPYC 9654 CPU.}
    \vspace{0.5em}
    \label{tab:appendix_software_version}
    \begin{tabular}{ll}
        \toprule
        Dependency & Version \\
        \midrule
        \multicolumn{2}{l}{\textit{Runtime and numerical libraries}} \\
        Python & 3.10.16 \\
        PyTorch & 2.9.1+cu126 \\
        Torchvision & 0.24.1+cu126 \\
        NumPy & 1.26.4 \\
        SciPy & 1.15.3 \\
        einops & 0.8.2 \\
        PyYAML & 6.0.3 \\
        Pillow & 11.3.0 \\
        plyfile & 1.1.3 \\
        OpenCV (opencv-python) & 4.11.0.86 \\
        LPIPS & 0.1.4 \\
        tqdm & 4.67.3 \\
        colorama & 0.4.6 \\
        \midrule
        \multicolumn{2}{l}{\textit{Compilation and CUDA extensions}} \\
        CUDA Toolkit (nvcc) & 12.6.77 \\
        GCC/G++ & 12.3.0 \\
        setuptools & 80.10.2 \\
        Ninja & 1.13.0 \\
        CUTLASS* & \texttt{291300ffffa3} \\
        GLM & 0.9.9.9 \\
        \bottomrule
    \end{tabular}
    \par\smallskip
    \begin{minipage}{0.95\linewidth}
        \footnotesize
        *CUTLASS is identified by its Git commit.
    \end{minipage}
\end{table}

\end{document}